\documentclass[letterpaper,10pt]{article}
\usepackage{cite}

\usepackage[pdftex]{graphicx}

\usepackage{amsfonts,amsmath,amsthm,amsxtra,amssymb,mathrsfs,dsfont}
\usepackage{mathtools}
\usepackage{array}
\usepackage[margin=1.2in]{geometry}

\usepackage{subfig}
\usepackage{hyperref}

\title{(Quasi)Periodicity Quantification in Video Data, Using Topology}

\author{Christopher~J.~Tralie$^\dag$
        and~Jose~A.~Perea*
\thanks{$\dag$ Department of Electrical And Computer Engineering,
         Duke University,Durham,
         NC, USA. e-mail: ctralie@alumni.princeton.edu}% <-this % stops a space
\thanks{* Department of Mathematics and Department of Computational Mathematics, Science \& Engineering,
         Michigan State University, East Lansing,
         MI, USA.  e-mail: joperea@math.msu.edu }% <-this % stops a space
}

\begin{document}


\maketitle

% REQUIRED
\begin{abstract}
This work introduces a novel framework for quantifying the presence and strength of
recurrent dynamics in video data. Specifically, we provide continuous measures of periodicity (perfect repetition) and quasiperiodicity (superposition of periodic modes with non-commensurate periods), in a way which does not require segmentation, training, object tracking or 1-dimensional surrogate signals.
Our methodology operates directly on  video data. The approach combines ideas from nonlinear time series analysis (delay embeddings) and computational topology (persistent homology), by translating the problem of finding recurrent dynamics in video data, into the problem of determining the circularity or toroidality of an associated geometric space. Through extensive testing, we show the robustness of our  scores with respect to several noise models/levels; we show that our periodicity score is superior to other methods when compared to human-generated periodicity rankings; and furthermore, we show that our quasiperiodicity score clearly indicates the presence of biphonation in videos of vibrating vocal folds, which has never before been accomplished end to end quantitatively.
\end{abstract}

%\begin{document}

\section{Introduction}
Periodicity characterizes many natural motions including animal locomotion (walking/wing flapping/slithering), spinning wheels, oscillating pendulums, etc.  Quasiperiodicity, thought of as the superposition of  non-commensurate frequencies, occurs  naturally during transitions from ordinary to chaotic dynamics \cite{gollub1975onset}.  The goal of this work is to automate the analysis of videos capturing periodic and quasiperiodic motion. In order to identify both classes of motion in a unified framework, we generalize 1-dimensional (1D) sliding window embeddings \cite{takens1981detecting} to reconstruct periodic and quasiperiodic attractors from videos\footnote{Some of the analysis and results appeared as part of the Ph.D. thesis of the first author \cite{tralie2017geometric}.}\footnote{Code to replicate results: \url{https://github.com/ctralie/SlidingWindowVideoTDA}}\footnote{Supplementary material and videos: \url{https://www.ctralie.com/Research/SlidingWindowVideoQuasi}}.  We analyze the resulting attractors using  persistent homology, a technique which combines   geometry and topology (Section~\ref{sec:tda}), and we return scores in the range $[0, 1]$ that indicate the degree of periodicity or quasiperiodicity in the corresponding video.  We show that our periodicity measure compares favorable to others in the literature when ranking videos (Section~\ref{sec:ranking}).  Furthermore, to our knowledge, there is no other method able to quantify the existence of quasiperiodicity directly from video data.

Our approach is fundamentally different from most others which quantify periodicity in video.  For instance, it is common to derive 1D signals from the video and apply Fourier or autocorrelation to measure periodicity.  By contrast, our technique operates on raw pixels, avoiding common video preprocessing and tracking entirely.  Using geometry over Fourier/autocorrelation also has advantages for our applications.  In fact, as a simple synthetic example shows (Figure~\ref{fig:quasiperiodicPSD}), the Fourier Transform of quasiperiodic signals is often very close to the Fourier transform of periodic signals.  By contrast, the sliding window embeddings we design yield starkly different geometric structures in the periodic and quasiperiodic cases.  We exploit this to devise a quasiperiodicity measurement, which we use to indicate the degree of ``biphonation'' in videos of vibrating vocal folds (Section~\ref{sec:vocalcords}), which is useful in automatically diagnosing speech pathologies.

In the context of applied topology, our quasiperiodicity score is one of the first applications of persistent $H_2$ to high dimensional data, which is largely possible due to recent advancements in the computational feasibility of persistent homology \cite{Ripser}.

%Though our method could be used for classification, we focus in this work on developing dataset independent, continuous measures of periodicity and quasiperiodicity.

%\begin{itemize}
%    \item Describe problem
%    \item other approaches
%    \item We interpret a video as a time series of frames
%    \item Use delay embeddings from the dynamical systems and nonlinear time series analysis literature
%    \item The shape of these embeddings reflects properties of the underlying dynamics: periodicity, quasiperiodicity, etc.
%    \item We use tools from computational topology to quantify the shape of said embeddings and derive scores for periodic and quasiperiodic dynamic regimes
%    \item We test our method extensively and compare to other popular approaches in the literature.
%\end{itemize}

\subsection{Prior Work on Recurrence in Videos}
\label{sec:prior}
%
%We first review some of the existing work addressing  detection and quantification of periodicity in videos, primarily in the computer vision literature.
\subsubsection{1D Surrogate Signals}
One common strategy for detecting periodicity in video is to derive a 1D function  to act as a surrogate for its dynamics, and then to use either frequency domain (Fourier transform) or time domain (autocorrelation, peak finding) techniques.
One of the earliest works in this genre finds level set surfaces in a spatiotemporal ``XYT'' volume of video (all frames stacked on top of each other), and then uses curvature scale space on curves that live on these ``spatiotemporal surfaces'' as the 1D function \cite{allmen1990cyclic}.  \cite{polana1997detection} use Fourier Transforms on pixels which exhibit motion, and define a measure of periodicity based on the energy around the Fourier peak and its harmonics.  \cite{goldenberg2005behavior} extract contours and find eigenshapes from the contours to classify and parameterize motion within a period.  Frequency estimation is done by using Fourier analysis and peak detection on top of other 1D statistics derived from the contours, such as area and center of mass.  Finally, \cite{yang2016time} derive a 1D surrogate function based on mutual information between the first and subsequent frames, and then look for peaks in the similarity function with the help of a watershed method.

\subsubsection{Self-Similarity Matrices}
\label{sec:ssmbg}
Another class of techniques relies on self-similarity matrices (SSMs) between frames, where similarity can be defined in a variety of ways.  \cite{seitz1997view} track a set of points on a foreground object and compare them with an affine invariant similarity.  Another widely recognized technique for periodicity quantification \cite{cutler2000robust}, derives periodicity measures based on self-similarity matrices of L1 pixel differences.  This technique has inspired a diverse array of applications, including analyzing the cycles of expanding/contracting jellyfish \cite{plotnik2002quantification}, analyzing bat wings \cite{atanbori2013analysis}, and analyzing videos of autistic spectrum children performing characteristic repetitive motions such as ``hand flapping'' \cite{kumdee2015repetitive}.  We compare to this technique in Section~\ref{sec:ranking}.

\subsubsection{Miscellaneous Techniques for Periodic Video Quantification}
There are also a number of works that don't fall into the two categories above.  Some works focus solely on walking humans, since that is one of the most common types of periodic motion in videos of interest to people.  \cite{niyogi1994analyzing} look at the ``braiding patterns'' that occur in XYT slices of videos of walking people.  \cite{huang2016camera} perform blob tracking on the foreground of a walking person, and use the ratio of the second and first eigenvalues of PCA on that blob.

For more general periodic videos, \cite{wang2009quasi} make a codebook of visual words and look for repetitions within the resulting string.  \cite{levy2015live} take a deep learning approach to counting the number of periods that occur in a video segment.  They use a 3D convolutional neural network on spatially downsampled, non-sequential regions of interest, which are uniformly spaced in time, to estimate the length of the cycle. Finally, perhaps the most philosophically similar work to ours is the work of \cite{vejdemo2015cohomological}, who use cohomology to find maps of MOCAP data to the circle for parameterizing periodic motions, though this work does not   provide a way to quantify periodicity.

\subsubsection{Our Work}

We show  that geometry provides a natural way to quantify recurrence (i.e. periodicity and quasiperiodicity) in video, by measuring the shape of delay embeddings.
In particular, we propose several optimizations (section \ref{sec:SlidingWindowVideo}) which make this approach feasible.
The resulting  measure of {\em quasiperiodicity}, for which quantitative approaches are lacking, is used in section~\ref{sec:vocalcords} to  detect anomalies in high-speed videos of vibrating vocal folds.  Finally, in contrast to both frequency and time domain techniques, our method does not rely on the period length being an integer multiple of the sampling rate.

\section{Background}

\subsection{Delay Embeddings And Their Geometry}

Recurrence in video data can be captured via the geometry of delay embeddings; we describe this next.

\subsubsection{Video Delay Embeddings} We will regard a video
as a sequence of grayscale\footnote{For color videos we can treat each channel independently, yielding a vector in $\mathbb{R}^{W \times H \times 3}$.  In practice, there isn't much of a difference between color and grayscale embeddings in our framework for the videos we consider.} image frames indexed by the positive real numbers.
That is, given positive integers $W$ (width) and  $H$ (height),
a  video with $W\times H$ pixels is a  function
\[
\begin{array}{rccl}
X: \mathbb{R}^+ \longrightarrow   \mathbb{R}^{W\times H}
\end{array}
\]
In particular, a sequence of  images
$X_1,X_2,\ldots \in \mathbb{R}^{W\times H}$ sampled at
discrete times $t_1 < t_2 < \cdots $ yields one such function
via interpolation.
For an integer $d \geq 0$, known as the \emph{dimension}, a real number $\tau > 0$, known as the \emph{delay}, and
a video $X : \mathbb{R}^+ \longrightarrow
\mathbb{R}^{W\times H}$,
we define the \textbf{sliding window} (also referred to as time delay) \textbf{embedding} of $X$ -- with
parameters $d$ and $\tau$ -- at time $t\in \mathbb{R}^+$
as the vector

\begin{equation}
\label{eq:delayvideo}
SW_{d, \tau} X(t) = \left[ \begin{array}{c} X(t) \\ X(t + \tau) \\ \vdots \\ X(t + d\tau) \end{array} \right]
\in \mathbb{R}^{W\times H \times (d+1)}
\end{equation}

The subset of $\mathbb{R}^{W\times H\times (d+1)}$
resulting from varying $t$ will be referred to as the
sliding window embedding of $X$.
We remark that since the pixel measurement locations are fixed, the sliding window embedding is an ``Eulerian'' view into the dynamics of the video.  Note that delay embeddings are generally applied to 1D time series, which can be viewed as 1-pixel videos ($W = H = 1$) in our framework.
Hence equation (\ref{eq:delayvideo})
is essentially the concatenation of the delay embeddings of each individual pixel in the video into one large vector.
One of the main points we leverage in this paper
is the fact that the geometry of the sliding window embedding
carries fundamental information about the original video.
We explore this next.

\subsubsection{Geometry of 1-Pixel Video Delay Embeddings}
%In order to describe the geometry of sliding window embeddings of videos with arbitrary resolution $W\times H$,
%we start with two  simple examples in the
%case of 1D time series (1-pixel videos).
As a motivating example, consider the   \emph{harmonic} (i.e. periodic) signal
\begin{equation}
f_h(t) =  \cos \left(\frac{\pi}{5} t\right) +  \cos \left(\frac{\pi}{15} t \right)
\end{equation}
and the {\em quasiperiodic} signal
\begin{equation}
f_q(t) =  \cos \left(\frac{\pi}{5} t\right) +  \cos \left(\frac{1}{5} t \right)
\end{equation}
We refer to $f_h$  as harmonic because its constitutive  frequencies,
$\frac{1}{10}$ and $\frac{1}{30}$, are \emph{commensurate};
that is, they are linearly dependent
over the rational numbers $\mathbb{Q} \subset \mathbb{R}$.
By way of contrast, the underlying frequencies
of the  signal $f_q$, $\frac{1}{10} $ and $ \frac{1}{10\pi}$,
are linearly independent over $\mathbb{Q}$
and hence \emph{non-commensurate}.
We use the term quasiperiodicity, as in the non-linear
dynamics literature  \cite{kantz2004nonlinear}, to denote the
superposition of periodic processes whose
frequencies are non-commensurate.
This differs from  other definitions in the literature (e.g. \cite{wang2009quasi,robinson2016topological}) which regard quasiperiodic as any deviation from perfect repetition.

\begin{figure}[!h]
\centering
\includegraphics[width=0.53\columnwidth]{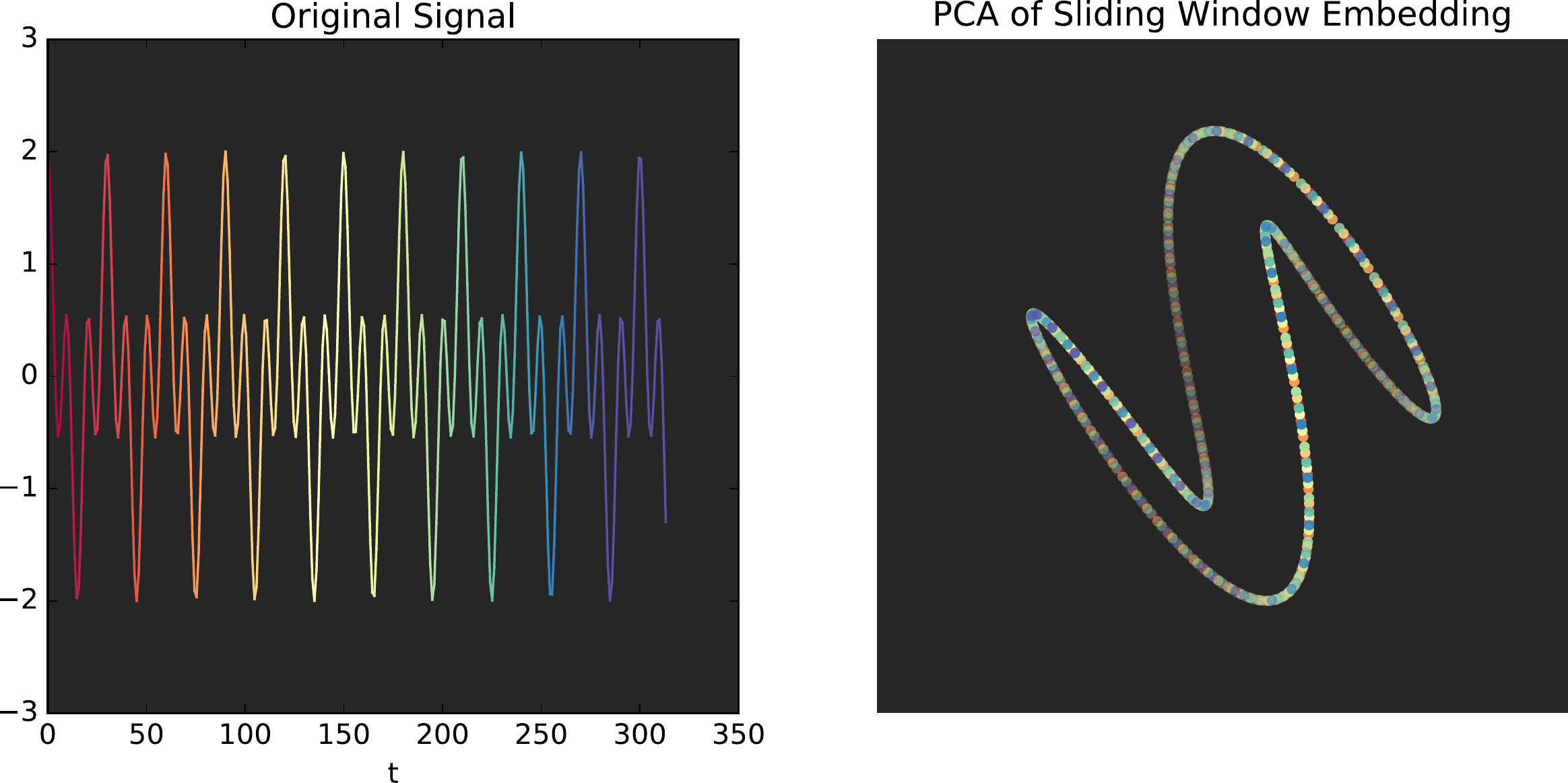}
\caption{Sliding window embedding of the harmonic signal $f_h$. Colors in the signal correspond to colors of the points in the PCA plot. The sliding window embedding traces a topological circle wrapped around a 2-dimensional torus.}
    \label{fig:Harmonic1D_NoPers}
\end{figure}

A geometric argument from  \cite{perea2015sliding} (see equation \ref{eq:SWtorus} below and the discussion that follows) shows that
given a periodic function $f : [0,2\pi] \longrightarrow \mathbb{R}$ with exactly $N$ harmonics,
if  $d \geq 2N$
and  $0 < \tau < \frac{2\pi}{d}$  then the sliding
window embedding $SW_{d,\tau}f$ is a
 topological circle
(i.e. a closed curve without self-intersections)
which wraps around an $N$-dimensional torus
\[
\mathbb{T}^N = \underbrace{S^1 \times \cdots \times S^1}_{N-times}
\;\;\; , \;\;\;
S^1 = \{z\in\mathbb{C} \; : \; |z|= 1\}
\]
As an illustration, we show in Figure \ref{fig:Harmonic1D_NoPers} a plot of $f_h$ and  of its sliding window embedding $SW_{d,\tau} f_h$, via
a PCA (Principal Component Analysis) 3-dimensional projection.

\begin{figure}[!h]
\centering
\includegraphics[width=0.53\columnwidth]{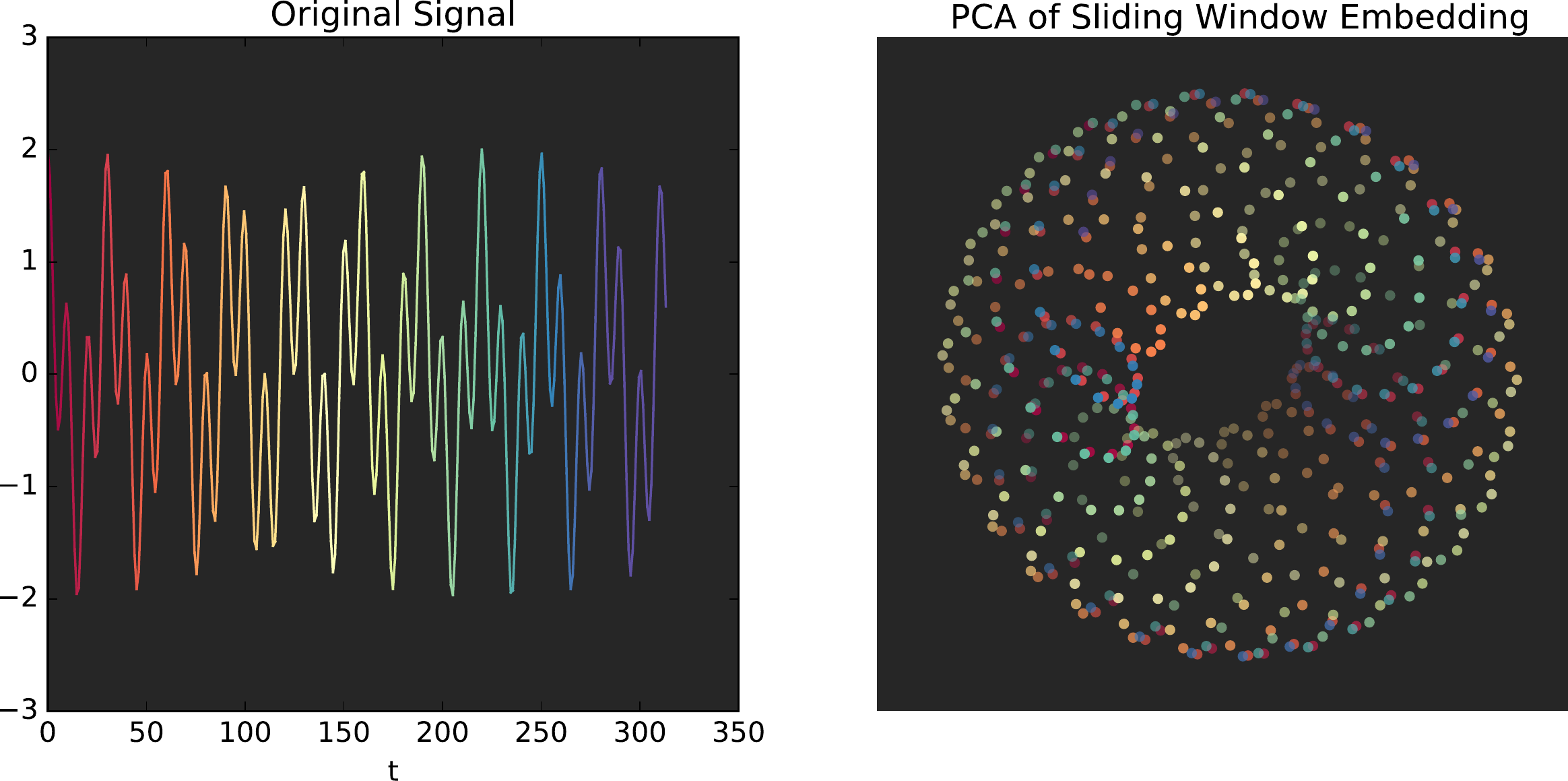}
\caption{Sliding window embedding of the quasiperiodic signal $f_q$. Colors in the signal correspond to colors of the points in the PCA plot. The sliding window embedding is dense in a 2-dimensional torus.}
    \label{fig:Quasiperiodic1D_NoPers}
\end{figure}

However, if $g: \mathbb{R}\longrightarrow \mathbb{R}$
is quasiperiodic with $N$ distinct non-commensurate
frequencies then, for appropriate $d$ and $\tau$,
$SW_{d,\tau}g$ is dense in (i.e. fills out) $\mathbb{T}^N$
\cite{perea2016persistent}.  Figure \ref{fig:Quasiperiodic1D_NoPers} shows a plot of the quasiperiodic signal $f_q(t)$ and a 3-dimensional projection, via PCA, of its sliding window embedding $SW_{d,\tau} f_q$.

\begin{figure}[!h]
	\centering	\includegraphics[width=0.53\columnwidth]{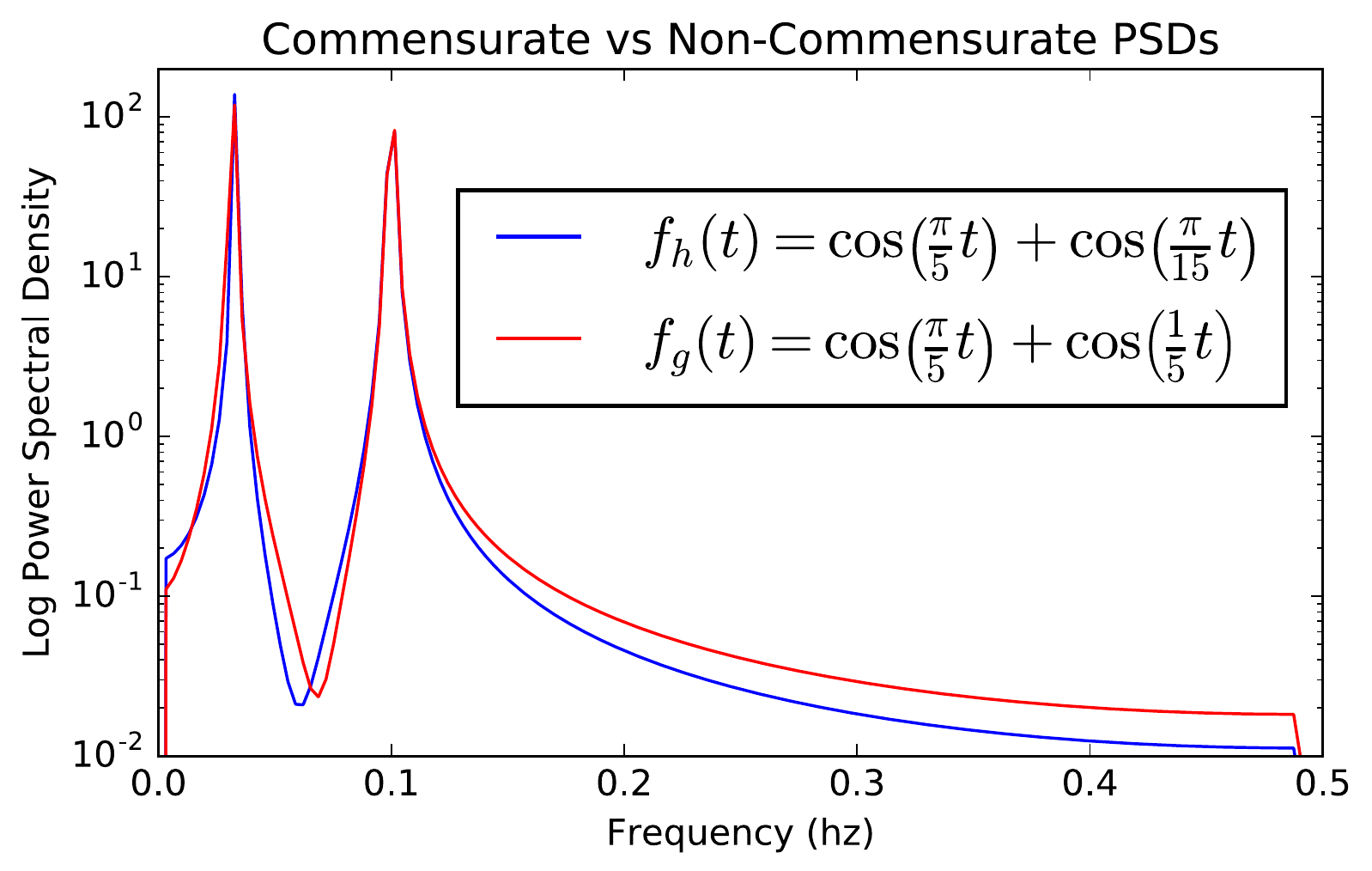}
	\caption{The power spectral densities (300 samples) of commensurate and non-commensurate signals with relative harmonics at ratios $3$ and $\pi$, respectively.  The difference is not nearly as evident as in the geometry of their sliding window embeddings.  Additionally, unless sampling is commensurate with a frequency, a fixed Fourier basis causes that frequency component to bleed into many frequency bins in a sinc-like pattern, making precise peak finding difficult.}
    \label{fig:quasiperiodicPSD}
\end{figure}

The difference in geometry of the delay embeddings is stark compared to the difference between their power spectral densities, as shown in Figure~\ref{fig:quasiperiodicPSD}.

Moreover, as we will see next, the  interpretation of periodicity and quasiperiodicity as circularity and toroidality of  sliding window embeddings remains true for videos with higher resolution (i.e. $\max\{W,H\} > 1$).
The rest of the paper will show how one can use \emph{persistent homology}, a tool from the field of computational topology, to quantify the presence of
(quasi)periodicity in a video by measuring the geometry
of its associated sliding window embedding.
In short, we propose a \emph{periodicity score} for a video $X$ which measures
the degree to which the sliding window embedding $SW_{d,\tau} X$ spans a topological circle, and a \emph{quasiperiodicity score} which quantifies the
degree to which $SW_{d,\tau} X$ covers a torus.
This approach will be validated extensively:
we show that our (quasi)periodicity detection method is robust under several
noise models (motion blur, additive
Gaussian white noise, and MPEG bit corruption);
we compare several periodicity quantification algorithms
and show that our approach is the most closely aligned with human subjects; finally, we provide an application to the automatic
classification of dynamic regimes in high-speed
laryngeal video-endoscopy.

\subsubsection{Geometry of Video Delay Embeddings}
Though it may seem daunting compared to the 1D case, the geometry of the delay embedding shares many similarities for periodic videos, as shown in \cite{tralie2016high}.
Let us argue why  sliding window embeddings from (quasi)periodic videos have the geometry we have described so far.
To this end, consider an example video $X$ that contains a set of $N$ frequencies $\omega_1, \omega_2, ..., \omega_N$.  Let the amplitude of the $n^{\text{th}}$ frequency and $i^{\text{th}}$ pixel be $a_{in}$.  For simplicity, but without loss of generality, assume that each is a cosine with  zero phase offset.  Then the time series at pixel $i$ can be written as
\begin{equation}
X_i(t) = \sum_{n = 1}^N a_{in}\cos(\omega_n t)
\end{equation}

\begin{figure}[!h]
	\centering
	\includegraphics[width=0.8\columnwidth]{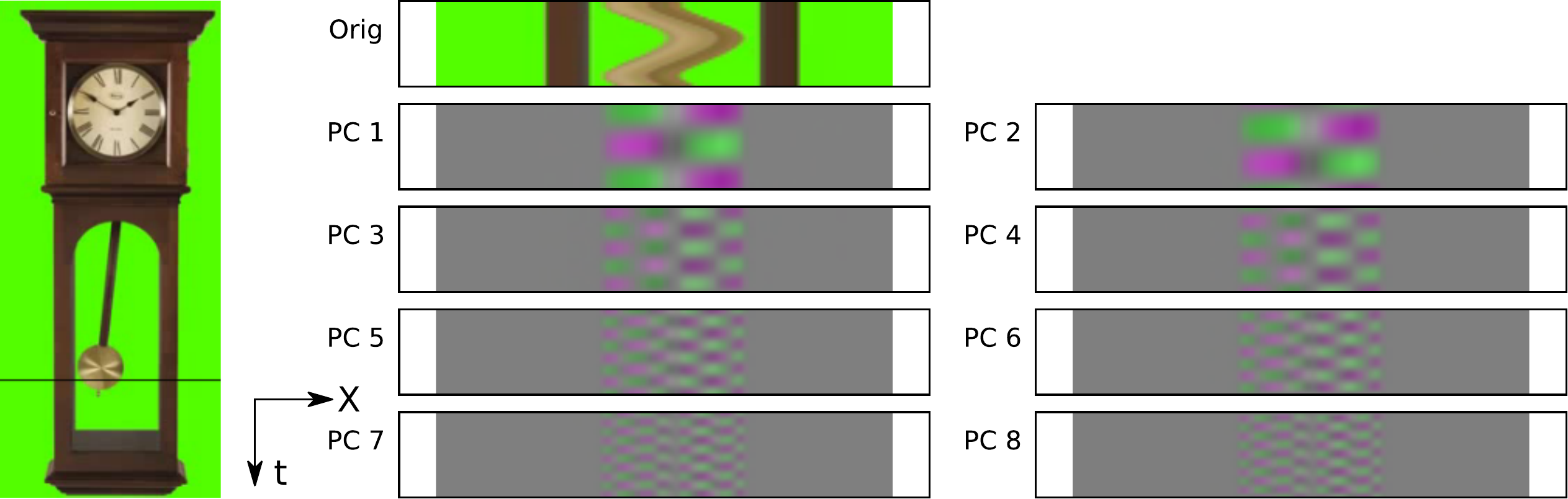}
	\caption{Showing an XT slice of the principal components on $SW_{d, 1}$ for the synthetic video of an oscillating pendulum; $d$ is chosen just under the period length ($\sim$ 25 frames).}
	\label{fig:VideoPCs}
\end{figure}

Grouping all of the coefficients together into a $(W\times H) \times N$ matrix $A$, we can write
\begin{equation}
X(t) = \sum_{n = 1}^N A^n cos(\omega_n t)
\end{equation}
where $A^n$ stands for the $n^{\text{th}}$ column of $A$.  Constructing a delay embedding as in Equation~\ref{eq:delayvideo}:
\begin{equation}
SW_{d, \tau}X(t) = \sum_{n = 1}^N \left[ \begin{array}{c}  A^n \cos(\omega_n t) \\ %A^n \cos(\omega_n (t + \tau)) \\
\vdots \\ A^n \cos(\omega_n (t + d \tau)  \end{array}  \right]
\end{equation}
and applying the cosine sum identity, we get
%
%\begin{equation}
%\begin{aligned}
%SW_{d, \tau}X(t) = \sum_{n = 1}^N \left[ \begin{array}{c}  A^n \\ A^n \cos(\omega_n \tau) \\ \vdots \\ A^n \cos(\omega_n d \tau)  \end{array}  \right] \cos(\omega_n t) - \\ \left[ \begin{array}{c}  0^n \\ A^n \sin(\omega_n \tau) \\ \vdots \\ A^n \sin(\omega_n d \tau)  \end{array}  \right] \sin(\omega_n t)
%\end{aligned}
%\label{eq:ellipseparam}
%\end{equation}
%
%Or written more succinctly
%
\begin{equation}\label{eq:SWtorus}
SW_{d, \tau}X(t) = \sum_{n = 1}^N \vec{u}_n\cos(\omega_n t) - \vec{v}_n \sin(\omega_n t)
\end{equation}
where $\vec{u}_n, \vec{v}_n \in \mathbb{R}^{  W \times H\times (d+1)}$
are constant vectors.
In other words, the sliding window embedding of this video is the sum of linearly independent ellipses, which lie in the space of
$d+1$ frame videos at resolution $W \times H$.
As shown in \cite{perea2015sliding} for the case of commensurate frequencies, when the window length is just under the length of the period, all of the $\vec{u}_n$ and $\vec{v}_n$ vectors become orthogonal, and so they can be recovered by doing PCA on $SW_{d, \tau}X(t)$.  Figure~\ref{fig:VideoPCs} shows the components of the first 8 PCA vectors for a horizontal line of pixels in a video of an oscillating pendulum. Note how the oscillations are present both temporally and spatially.
%, as suggested by Equation~\ref{eq:ellipseparam}.

\subsubsection{The High Dimensional Geometry of Repeated Pulses}

\begin{figure}[!htb]
	\centering
	\includegraphics[width=0.6\columnwidth]{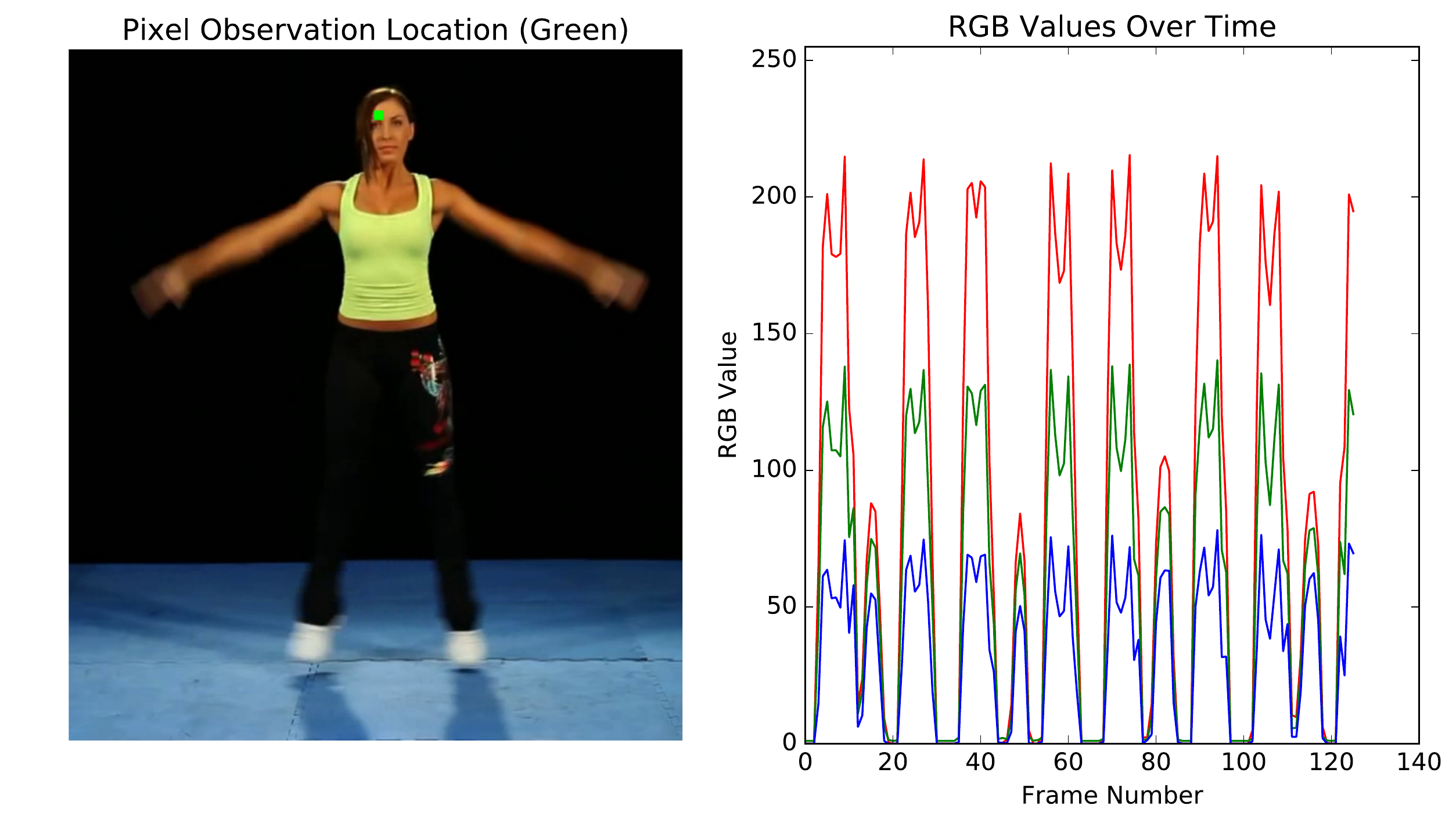}
	\caption{An example of an Eulerian pixel witnessing a foreground/background transition in a video of a woman doing jumping jacks.  Red, green, and blue channels are plotted over time.  These transitions induce a per pixel periodic signal with sharp transitions, which leads to high dimensionality in an appropriate sliding window embedding.}
	\label{fig:PixelFGBG}
\end{figure}

Using Eulerian coordinates has an important impact on the geometry of delay embeddings of natural videos.  As Figure~\ref{fig:PixelFGBG} shows, pixels often jump from foreground to background in a pattern similar to square waves.

These types of abrupt transitions require higher dimensional embeddings to reconstruct the geometry.  To see why, first extract one period of a signal with period $\ell$ at a pixel $X_i(t)$:
\begin{equation}
f_i(t) = \left\{ \begin{array}{cc} X_i(t) & 0 \leq t \leq \ell \\ 0 & \text{otherwise} \end{array} \right\}
\end{equation}
Then $X_i(t)$ can be rewritten in terms of the pulse as
\begin{equation}
X_i(t) = \sum_{m = -\infty}^{\infty} f_i(t - m\ell)
\end{equation}
Since $X_i(t)$ repeats itself, regardless of what $f_i(t)$ looks like, {\em periodic summation} discretizes the frequency domain \cite{pinsky2002introduction}
\begin{equation}
 \mathcal{F}\left\{  X_i(t) \right\}(k) \propto \sum_{m = -\infty}^{\infty} \mathcal{F}(f_i(t)) \left( \frac{m}{\ell} \right) \delta \left( \frac{m}{\ell} - k \right)
\end{equation}
Switching back to the time domain, we can write $X_i(t)$ as
\begin{equation}
\label{eq:harmonicsums}
X_i(t) \propto \sum_{m = -\infty}^{\infty} \mathcal{F} \left( f_i(t) \right) \left(\frac{m}{\ell} \right) e^{i \frac{2 \pi m}{\ell} t}
\end{equation}

In other words, each pixel is the sum of some constant offset plus a (possibly infinite) set of harmonics at integer multiples of $\frac{1}{\ell}$.  For instance, applying Equation~\ref{eq:harmonicsums} to a square wave of period $\ell$ centered at the origin is a roundabout way of deriving the Fourier Series
\begin{equation}
\sin \left( \frac{2 \pi}{\ell} t \right) + \frac{1}{3} \sin \left( \frac{6 \pi}{\ell} t \right) + \frac{1}{5} \sin \left( \frac{10 \pi}{\ell} t \right) + \hdots
\end{equation}
by sampling the sinc function $\sin(\pi \ell f)/(\pi f)$ at intervals of $m/2\ell$ (every odd $m$ coincides with $\pi/2 + k \pi$, proportional to $1/k$, and every even harmonic is zero conciding with $\pi k$).  In general, the sharper the transitions are in $X_i(t)$, the longer the tail of $\mathcal{F}\{f_i(t)\}$ will be, and the more high frequency harmonics will exist in the embedding, calling for a higher delay dimension to fully capture the geometry, since every harmonic lives on a linearly independent ellipse.  Similar observations about harmonics have been made in images for collections of patches around sharp edges (\cite{yu2012solving}, Figure 2).

\subsection{Persistent Homology}
\label{sec:tda}
Informally, topology is the study of properties of spaces which do not change after stretching without gluing or tearing.
For instance, the
number of connected components  and the number
of  (essentially different) 1-dimensional loops which do not bound
a 2-dimensional disk, are both topological properties of a space.
It follows that a circle and a square are topologically equivalent since one can deform one onto the other, but  a circle and a line segment are not because that would require either gluing the endpoints of the line segment or tearing the circle.
Homology \cite{hatcher2002algebraic} is a tool from algebraic topology designed to
measure these types of properties, and
persistent homology \cite{zomorodian05computing} is an adaptation of these ideas
to discrete collections of points (e.g., sliding window embeddings).
We briefly introduce these concepts next.

\subsubsection{Simplicial Complexes}
A simplicial complex is a
combinatorial object used to represent and discretize a continuous space.
With a discretization available, one can then
compute topological properties
by algorithmic means.
Formally,
a \textbf{simplicial complex}
with vertices in a nonempty set $V$ is a collection $K$
of nonempty finite subsets $\sigma \subset V$
so that $\emptyset \neq \tau \subset \sigma \in K$
always implies $\tau \in K$.
An element $\sigma \in K$ is called
a \textbf{simplex}, and if  $\sigma$ has $(n+1)$
elements then   it is called an $n$-simplex.
The cases $n=0,1,2$ are special,  0-simplices are called vertices,
1-simplices are called edges and 2-simplices are called faces.
Here is an example to keep in mind: the circle $S^1 = \{z\in \mathbb{C} : |z| =1 \}$ is a continuous
space but its topology can be captured by a simplicial complex $K$
with three vertices $a,b,c$, and three edges $\{a,b\}$,  $\{b,c\}$, $\{a,c\}$.
That is, in terms of topological properties, the simplicial complex
\[K =\big \{\{a\},\{b\},\{c\}, \{a,b\}, \{b,c\}, \{a,c\}\big\}\]
can be regarded as a combinatorial surrogate for  $S^1$: they both have 1 connected component,  one 1-dimensional loop
which does not bound a 2-dimensional region, and no other features in higher dimensions.

\subsubsection{Persistent Homology of Point Clouds}
The sliding window embedding of a video $X$ is, in practice,  a finite
set $\mathbb{SW}_{d,\tau}X = \{SW_{d,\tau}X(t)  :  t\in T\}$
determined by a choice of $T\subset \mathbb{R}$  finite.
Moreover, since $\mathbb{SW}_{d,\tau}X \subset \mathbb{R}^{W\times H \times (d+1)}$
then the restriction of the ambient Euclidean distance endows $\mathbb{SW}_{d,\tau}X$ with the structure of a finite metric space.
Discrete metric spaces, also referred to as \emph{point clouds},
are trivial from a topological point of view:
a point cloud with $N$  points simply has $N$ connected components and no other features (e.g., holes) in higher dimensions.
However, when a point cloud has been sampled from/around a continuous
space with non-trivial topology (e.g., a circle or a torus),
one would expect that appropriate simplicial complexes with vertices
on the point cloud should reflect the topology of the
underlying continuous space.
This is what we will exploit next.

\begin{figure}[!htb]
  \centering
  \includegraphics[width=0.75\textwidth]{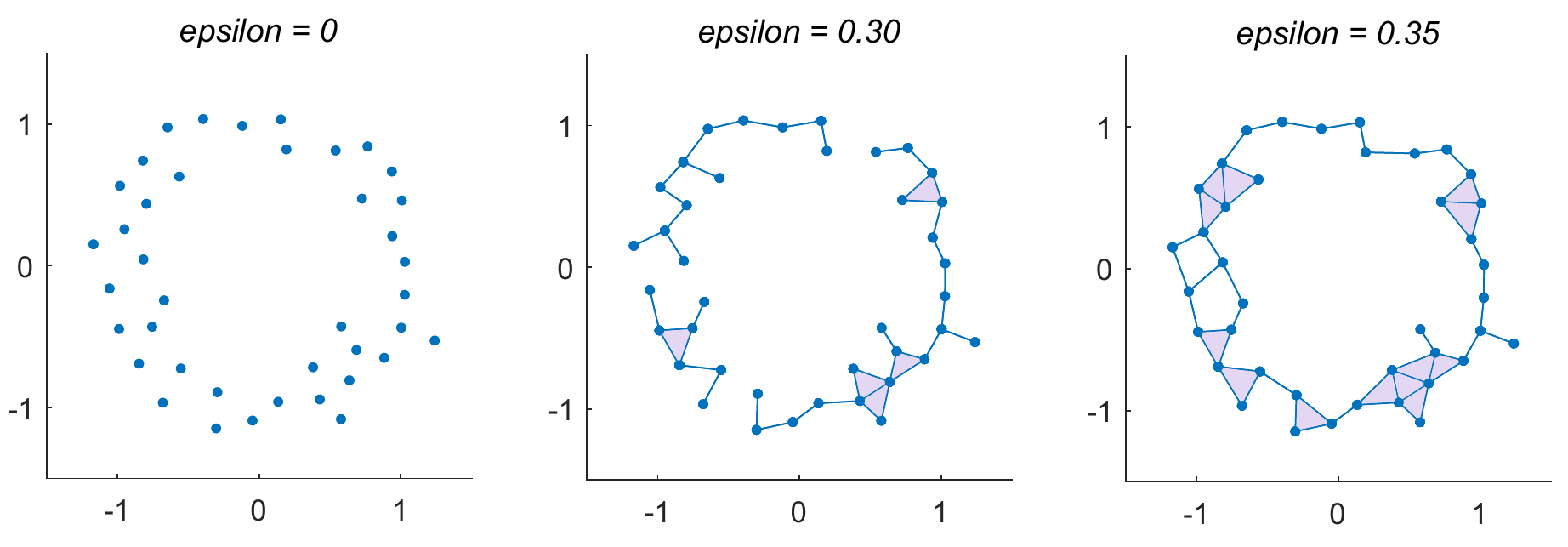}
  \caption{The Rips complex, at three different scales ($\epsilon = 0,\, 0.30,\, 0.35 $), on a point cloud with 40 points sampled around $S^1\subset \mathbb{R}^2$.}\label{fig:RipsFiltration}
\end{figure}

Given a point cloud $(\mathbb{X}, d_\mathbb{X})$  --
where $\mathbb{X}$ is a finite set and $d_\mathbb{X}: \mathbb{X} \times \mathbb{X} \longrightarrow [0,\infty)$ is a distance function --
the \emph{Vietoris-Rips complex} (or Rips complex for short) at scale
$\epsilon\geq0$ is the collection
of non-empty  subsets of $\mathbb{X}$ with diameter less than or equal to $\epsilon$:
\begin{equation}
R_\epsilon(\mathbb{X}) :=
\big\{ \sigma \subset \mathbb{X} : d_\mathbb{X}(x_1, x_2) \leq \epsilon, \; \forall x_i, x_j \in \sigma \big\}
\end{equation}
That is, $R_\epsilon(\mathbb{X})$ is the simplicial complex with vertex set equal to $\mathbb{X}$, constructed by adding 
an edge between any two vertices which are at most $\epsilon$ apart,
 adding all 2-dimensional triangular faces (i.e. 2-simplices) whose bounding edges are present, and
more generally, adding all the $k$-simplices whose $(k-1)$-dimensional bounding facets have been included.
We show in Figure \ref{fig:RipsFiltration} the evolution of the Rips complex on a set of points sampled around the unit circle.

The idea behind persistent homology is to track the evolution of topological features of complexes such as 
$R_\epsilon(\mathbb{X})$, as the scale parameter $\epsilon$ ranges from $0$ to some maximum value $\epsilon_\mathsf{max} \leq \infty$.
For instance, in Figure \ref{fig:RipsFiltration} one can see that $R_0 (\mathbb{X}) = \mathbb{X}$ has 40 distinct connect components (one for each point),
$R_{0.30}(\mathbb{X})$ has three connected components  and $R_{0.35}(\mathbb{X})$ has only one connected component; 
 this will continue to be the case for every $\epsilon \geq 0.35$.
Similarly, there are no closed loops in $R_0(\mathbb{X})$ or $R_{0.30}(\mathbb{X})$ bounding  empty regions, but this changes 
when  $\epsilon $ increases  to $0.35$. Indeed, $R_{0.35}(\mathbb{X})$ has three 1-dimensional holes: 
the central prominent hole, and the two small  ones to the left side. 
Notice, however, that as $\epsilon$ increases beyond $0.35$ these holes will be filled by the addition of new simplices; 
in particular, for $\epsilon > 2$
one has that $R_\epsilon(\mathbb{X})$ will have only one connected component and no other topological features in higher dimensions.

\begin{figure}[!htb]
	\centering	\includegraphics[width=0.7\columnwidth]{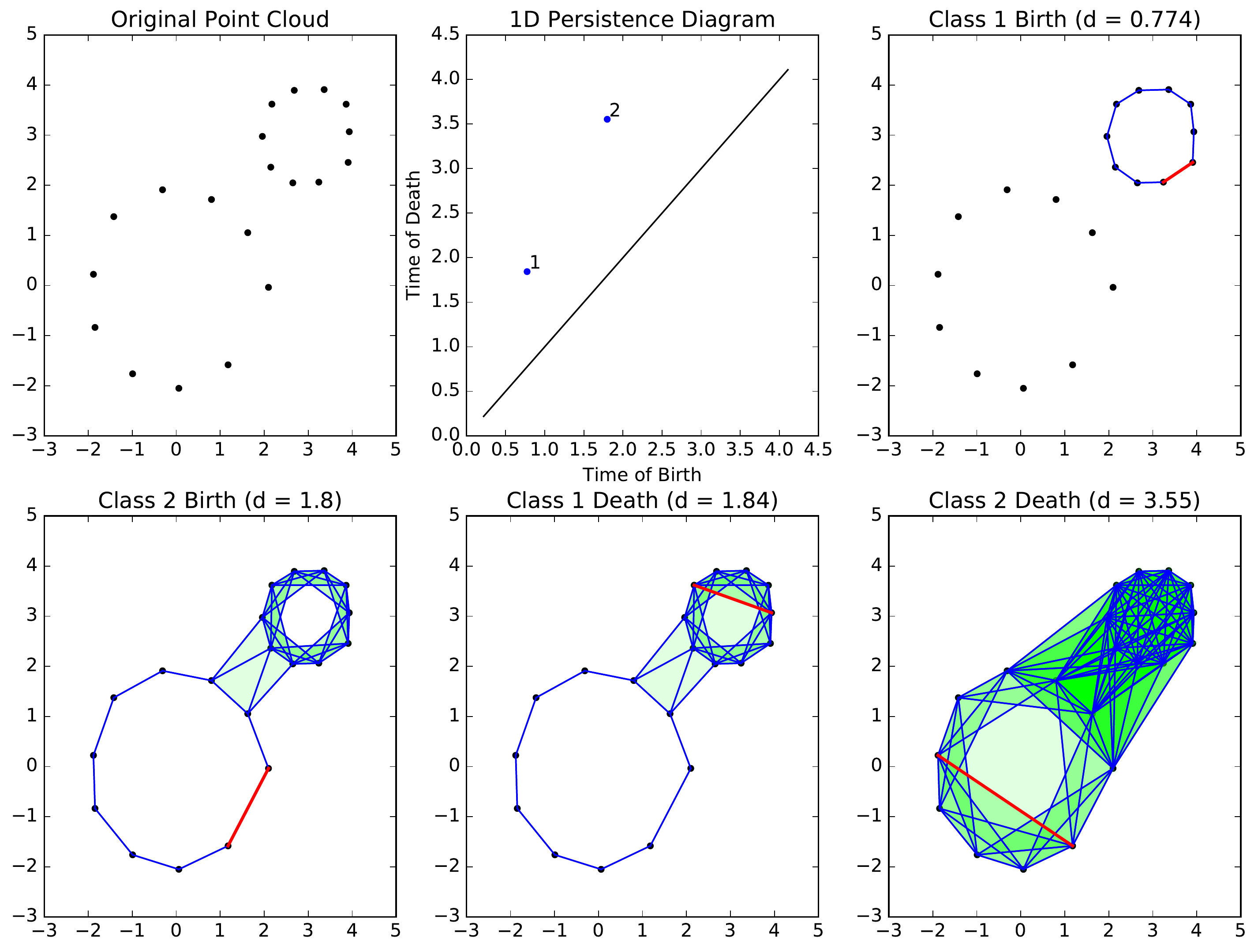}
	\caption{From a point cloud to the 1-dimensional persistence diagram of its Rips filtration.  Connected edges in the Rips filtration are drawn in blue, the birth/death of a class is indicated in red,  and filled in triangles are shaded green.}
	\label{fig:PersistenceExample1D}
\end{figure}

The family 
$\mathcal{R}(\mathbb{X}) = \{R_\epsilon(\mathbb{X})\}_{\epsilon\geq 0}$
is  known as the
\emph{Rips filtration} of $\mathbb{X}$, and the emergence/dissapearence of topological 
features in each dimension (i.e., connected components, holes, voids, etc) as $\epsilon$ changes, can be codified 
in what are  referred to as the \textbf{persistence diagrams} of $\mathcal{R}(\mathbb{X})$.
Specifically, for each dimension $n = 0,1,\ldots$ (0 = connected components, 1 = holes, 2 = voids, etc)
one can record the value of $\epsilon$ for which a particular $n$-dimensional topological feature of the Rips filtration appears (i.e. its birth time),
and when it disappears (i.e. its death time). 
The birth-death times $(b,d)\in \mathbb{R}^2$ of $n$-dimensional features for $\mathcal{R}(\mathbb{X})$ form a multiset $\mathsf{dgm}_n(\mathcal{R}(\mathbb{X}))$ ---
i.e. a set whose elements can come with repetition --- known as the $n$-dimensional persistence diagram of the Rips filtration on $\mathbb{X}$.
Since $\mathsf{dgm}_n(\mathcal{R}(\mathbb{X}))$ is just a collection of points
in the region $\{(x,y) \in \mathbb{R}^2 : 0 \leq x < y\}$,  we will visualize it as  a scatter plot.
The \emph{persistence} of a topological feature with birth-death times  $(b,d)$ is the quantity $d-b$, i.e. its lifetime.
We will also include the diagonal $y= x$  in the scatter plot in order to visually
convey the persistence  of each  birth-death
pair.
In this setting, points far from the diagonal (i.e. with large persistence)
represent topological features which are stable across scales and hence
deemed significant,
while points near the diagonal (i.e. with small persistence)
are often associated with unstable features.
We illustrate in Figure \ref{fig:PersistenceExample1D}
the process of going from a point cloud to
the 1-dimensional persistence diagram of its Rips filtration.

We remark that the computational task of determining all non-equivalent
persistent homology classes of a filtered simplicial complex can, surprisingly, be reduced to computing the homology of a single
simplicial complex \cite{zomorodian05computing, Ripser}.
This  is in fact a problem in linear algebra that can be
solved via elementary row and column operations on appropriate
boundary matrices.

\begin{figure}[!htb]
\centering
\includegraphics[width=0.53\columnwidth]{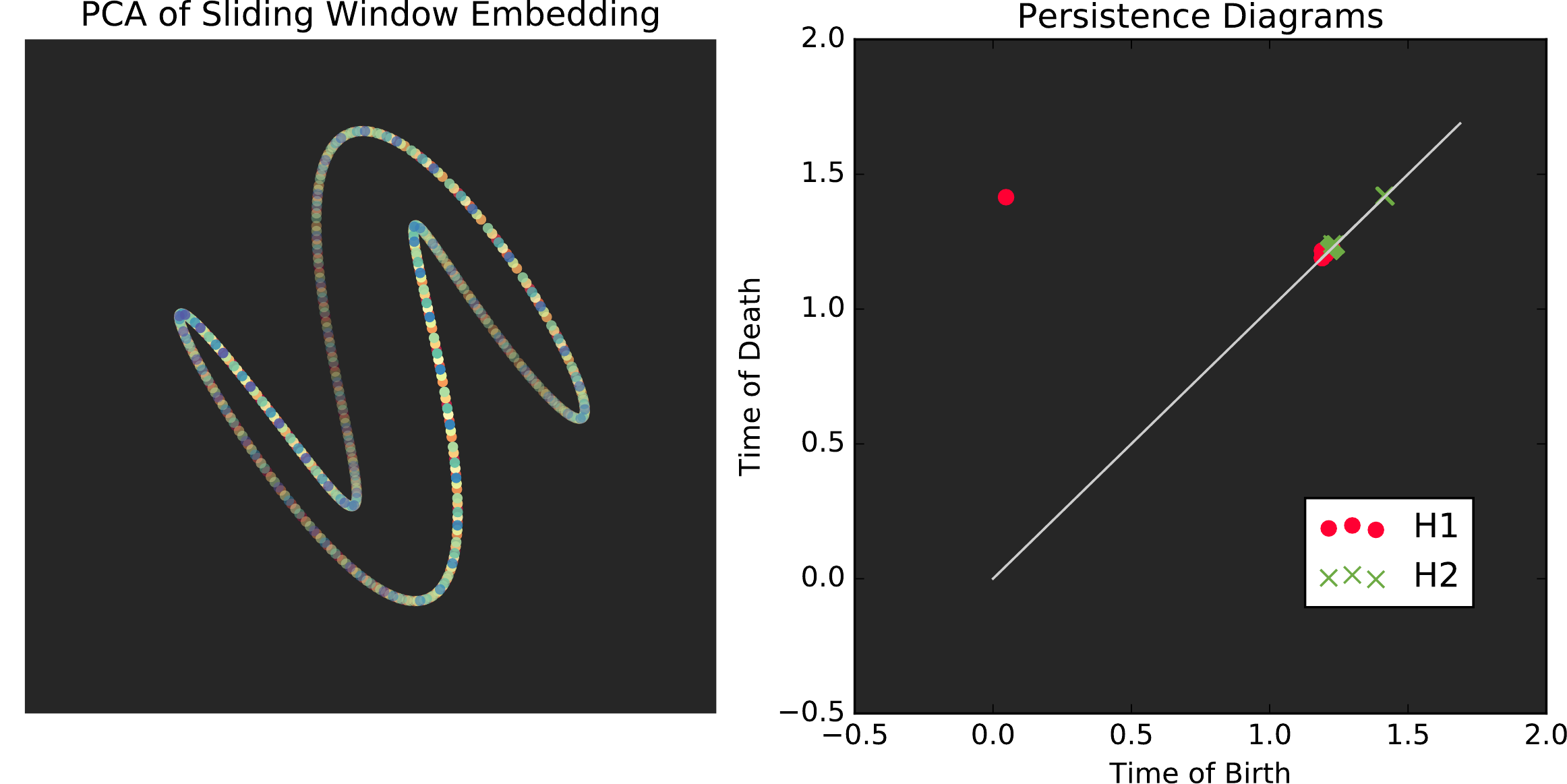}
\caption{Sliding window embedding of the harmonic signal $f_h$ (left) and the $n$-dimensional persistence diagrams $n=1,2$ (right) of the associated Rips filtration. The sliding window embedding $SW_{d,\tau} f_h $ traces a topological circle wrapped around a 2-dimensional torus. The persistence diagram in dimension one ($H_1$) shows only one birth-death pair with prominent persistence; this is consistent with a point cloud sampled around a space with the topology of a circle.}
    \label{fig:Harmonic1D}
\end{figure}

\begin{figure}[!htb]
\centering
\includegraphics[width=0.53\columnwidth]{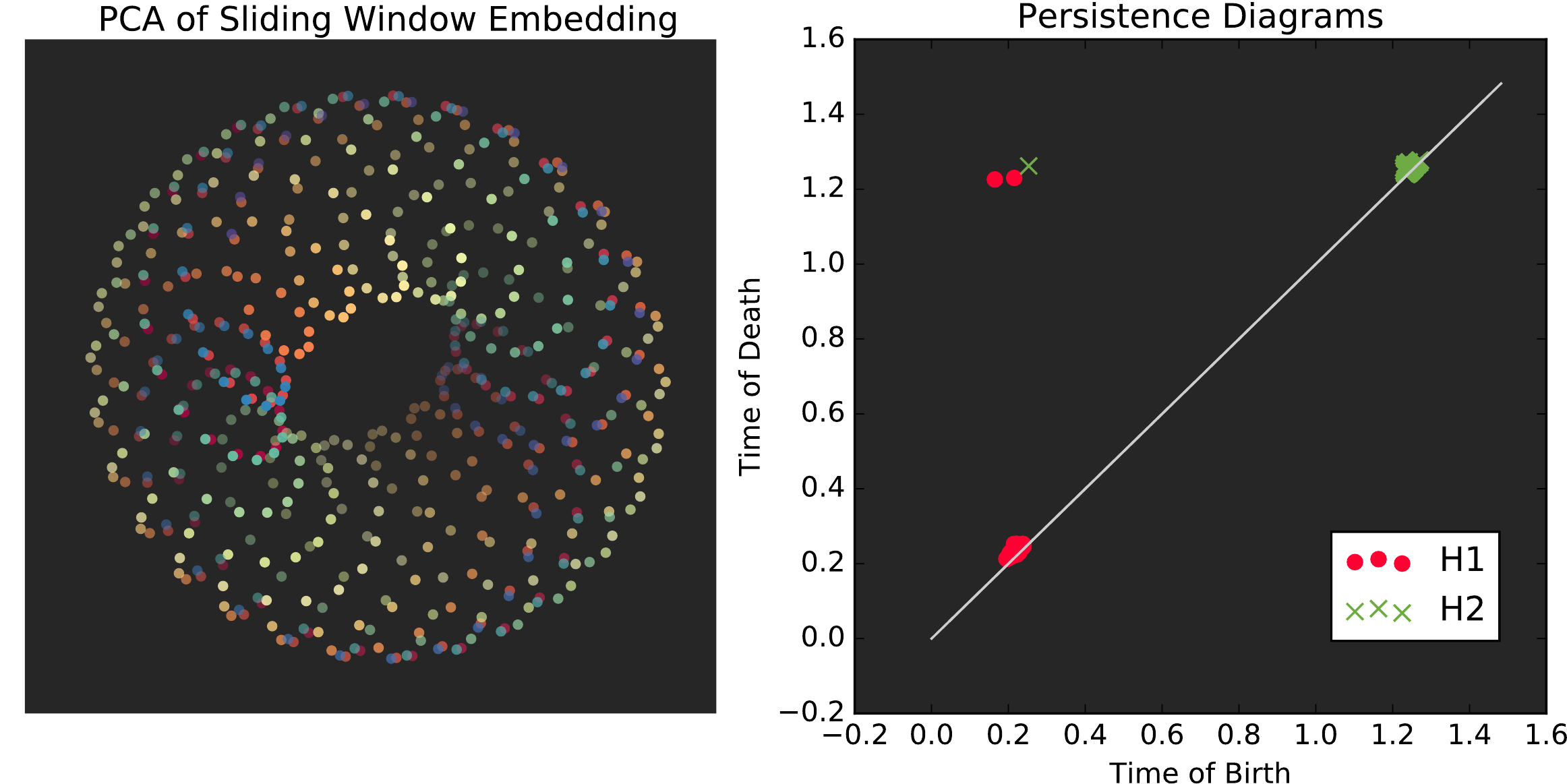}
\caption{Sliding window embedding of the quasiperiodic signal $f_q$ (left) and the $n$-dimensional persistence diagrams $n=1,2$ (right) of the associated Rips filtration. The sliding window embedding $SW_{d,\tau} f_q$ is dense on a 2-dimensional torus. The persistence diagram in dimension one ($H_1$) shows two birth-death pairs with prominent persistence, while the persistence diagram in dimension two ($H_2$) shows one prominent birth-death pair; this is consistent with a point cloud sampled around a space with the topology of a 2-dimensional torus.}
    \label{fig:Quasiperiodic1D}
\end{figure}

The persistent homology of  $\mathcal{R}\big(\mathbb{SW}_{d,\tau}X\big)$,
and in particular its $n$-dimensional persistence diagrams for $n=1,2$,
are the objects we will use to quantify periodicity and quasiperiodicity in a video $X$.
Figures \ref{fig:Harmonic1D} and
\ref{fig:Quasiperiodic1D}
show the persistence diagrams of the Rips filtrations, on the sliding window embeddings,
for the commensurate and non-commensurate signals from Figures~\ref{fig:Harmonic1D_NoPers} and \ref{fig:Quasiperiodic1D_NoPers}, respectively.  We use fast new code from the ``Ripser'' software package to make persistent $H_2$ computation feasible \cite{Ripser}.

%
%\begin{figure}
%	\centering	   \includegraphics[width=\columnwidth]{TorusPersistenceExample.pdf}
%    \caption{500 points are sampled on the torus.  Two 1 cycles and one 2 cycle are evident in the persistence diagram.}
%	\label{fig:TorusPersistenceExample}
%\end{figure}
%

\section{Implementation Details}
\label{sec:SlidingWindowVideo}

%
%
%Now that we have set the stage for our work, we are ready to discuss some of the computational aspects of sliding window videos in more detail.
%

\subsection{Reducing Memory Requirements with SVD}
\label{sec:svdmemory}
Suppose we have a video which has been discretely sampled at $N$ different frames at a resolution of $W \times H$, and we do a delay embedding with dimension $d$, for some arbitrary $\tau$.  Assuming 32 bit floats per grayscale value, storing the sliding window embedding requires $4WHN(d+1)$ bytes.  For a low resolution $200 \times 200$ video only 10 seconds long at 30fps, using $d = 30$  already exceeds 1GB of memory.
In what follows we will address the memory requirements and ensuing computational burden to construct and access the sliding window embedding.
Indeed, constructing the Rips filtration  only requires  pairwise distances between different delay vectors, this enables a few optimizations.

First of all, for $N$ points in $\mathbb{R}^{WH}$, where $N \ll WH$, there exists an $N$-dimensional linear subspace which contains them.
In particular let $A$ be the $(W \times H) \times N$ matrix with each video frame along a column.
Performing a singular value decomposition $A = USV^T$, yields
a matrix  $U$ whose columns form an orthonormal basis for the aforementioned $N$-dimensional linear subspace.
Hence, by finding the coordinates of the original frame vectors with respect to this orthogonal basis
\begin{equation}
\hat{A} = U^TA = U^TUSV = SV
\end{equation}
and using the coordinates of the columns of $SV$ instead of the original pixels,
we get a sliding window embedding of lower dimension
\begin{equation}
\label{eq:delayvideopca}
SW_{d, \tau} (t)= \left[ \begin{array}{c} U^TX(t) \\ \vdots \\ U^TX(t + d\tau) \end{array} \right]
\end{equation}
for which
\[
\|SW_{d,\tau}(t) - SW_{d,\tau}(t')\| =
\|SW_{d,\tau}X(t) - SW_{d,\tau}X(t')\|
\]

Note that $SV$ can be computed by finding the eigenvectors of $A^TA$;
this has a cost of $O(W^2H^2 + N^3)$ which is dominated by $W^2H^2$ if $WH \gg N$.  In our example above, this alone reduces the memory requirements from 1GB to 10MB.  Of course, this procedure is the most effective for short videos where there are actually many fewer frames than pixels, but this encompasses most of the examples in this work.  In fact, the break-even point for a 200x200 30fps video is 22 minutes.  A similar approach was used in the classical work on Eigenfaces \cite{turk1991eigenfaces} when computing the principal components over a set of face images.

\subsection{Distance Computation via Diagonal Convolutions}
\label{sec:diagconv}

A different optimization is possible if $\tau = 1$; that is, if delays are taken exactly on frames and no interpolation is needed.  In this case, the squared Euclidean distance between $SW_{d, 1}X(i)$ and $SW_{d, 1}X(j)$ is
\begin{equation}
\label{eq:diagconv}
||SW_{d, 1}X(i) - SW_{d, 1}X(j)||_2^2 = \sum_{m = 0}^d ||X(i+m) - X(j+m)||_2^2
\end{equation}

Let $D^2_X$ be the $N \times N$ matrix of all pairwise squared Euclidean distances between frames (possibly computed with the memory optimization in Section~\ref{sec:svdmemory}), and let $D^2_Y$ be the $(N-d) \times (N-d)$ matrix of all pairwise distances between delay frames.
Then Equation~\ref{eq:diagconv} implies that $D^2_Y$ can be obtained from $D^2_X$ via convolution with a ``rect function'', or a vector of $1$s of length $d+1$, over all diagonals in $D^2_X$ (i.e. a moving average).  This can be implemented in time $O(N^2)$ with cumulative sums.
Hence,  regardless of how $d$ is chosen, the computation and memory requirements for computing $D^2_Y$ depend only on the number of frames in the video.  Also, $D_Y$ can simply be computed by taking the entry wise square root of $D^2_Y$, another $O(N^2)$ computation.  A similar scheme was used in \cite{huang2010shape} when comparing distances of 3D shape descriptors in videos of 3D meshes.

\begin{figure}[!htb]
	\centering	\includegraphics[width=0.53\columnwidth]{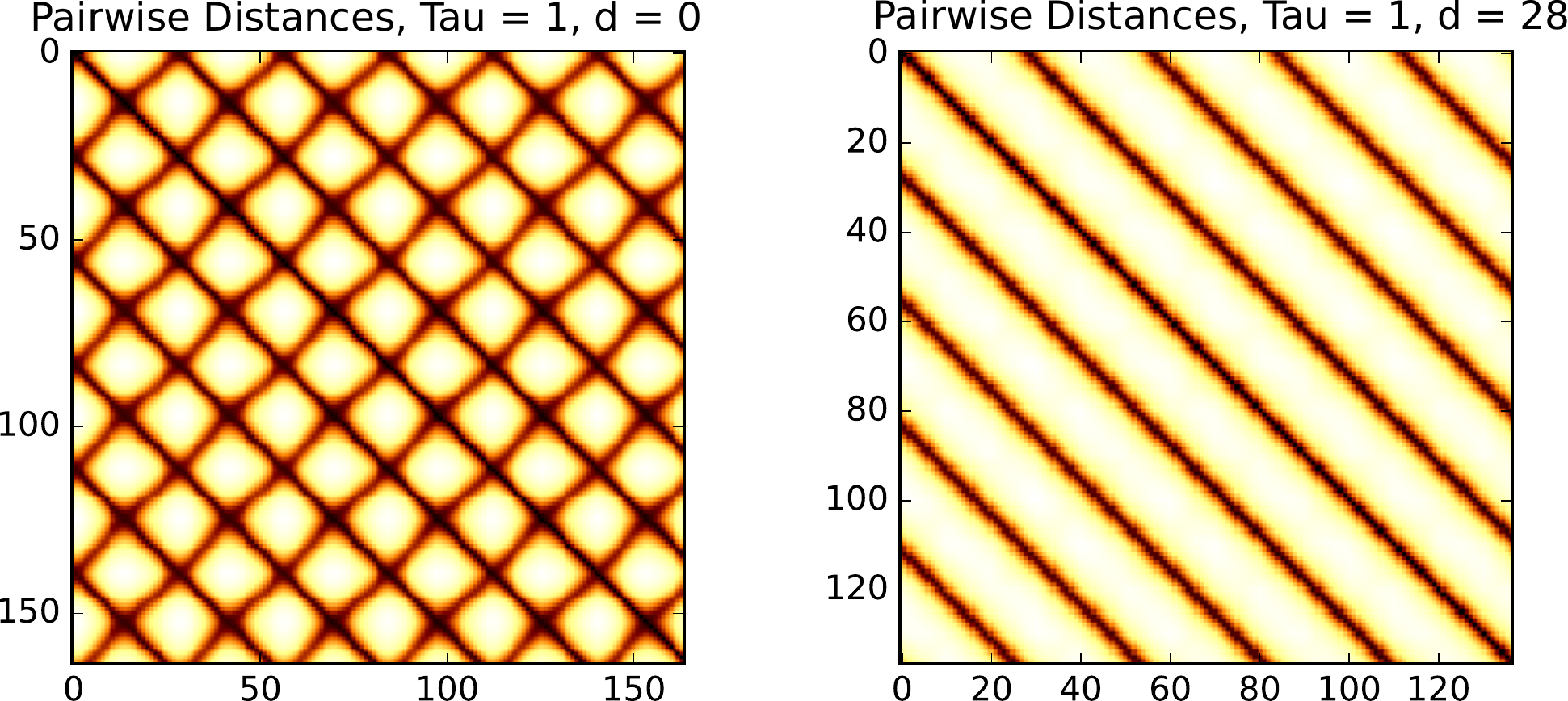}
	\caption{Self-similarity matrices $D_{Y_{0, 1}}$ and $D_{Y_{28, 1}}$ for a video of the oscillating pendulum.  Bright colors indicate far distances and dark colors indicate near distances.  This example clearly shows how adding a delay embedding is like performing block averaging along all diagonals of the pairwise distance matrices, and it gets rid of the mirror symmetry.}
	\label{fig:PendulumDelayNoDelay}
\end{figure}

Figure~\ref{fig:PendulumDelayNoDelay} shows self-similarity matrices on embeddings of the pendulum video with no delay and with a delay approximately matching the period.
The effect of a moving average along diagonals with delay eliminates the anti-diagonals caused by the video's mirror symmetry.

\begin{figure}[!htb]
	\centering
	\includegraphics[width=0.8\columnwidth]{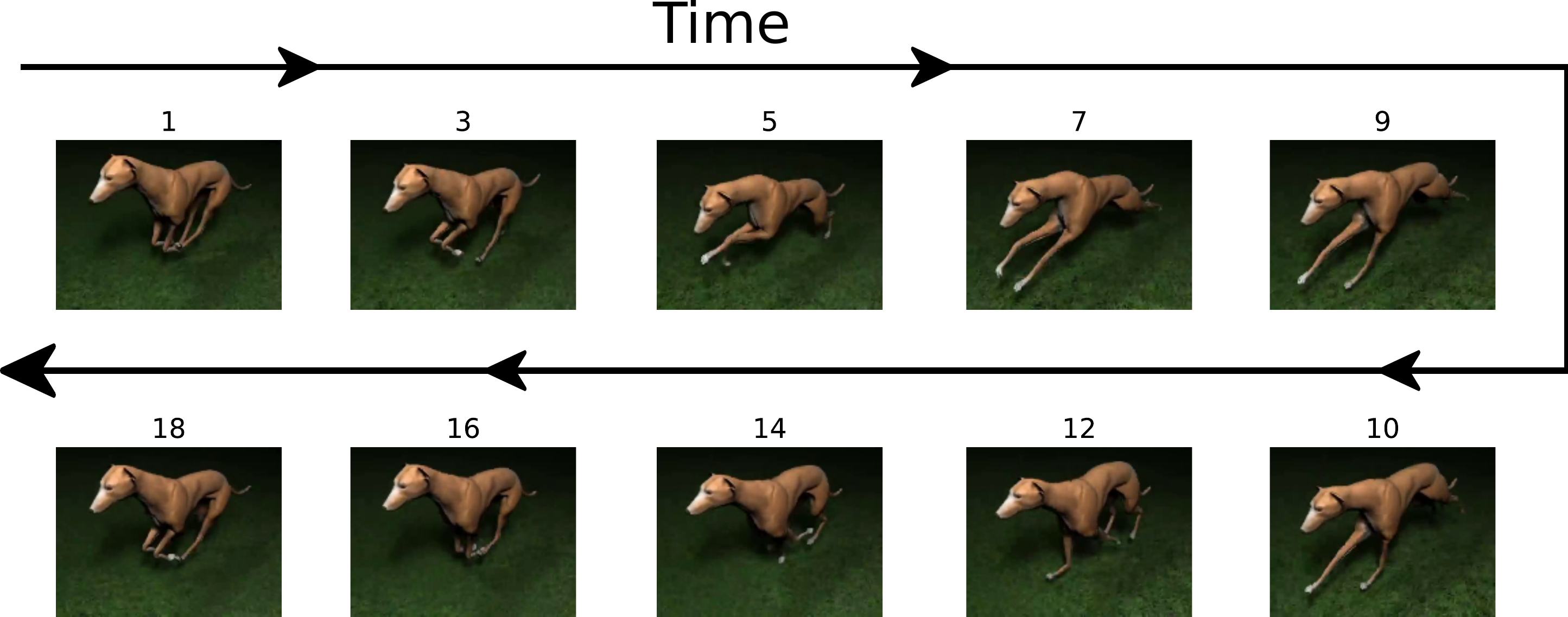}
	\caption{An animation of a periodic video of a running dog, which, unlike an oscillating pendulum, does not have mirror symmetry in the second half of its period.}
	\label{fig:DogVideo}
\end{figure}

\begin{figure}[!htb]
	\centering
	\includegraphics[width=0.53\columnwidth]{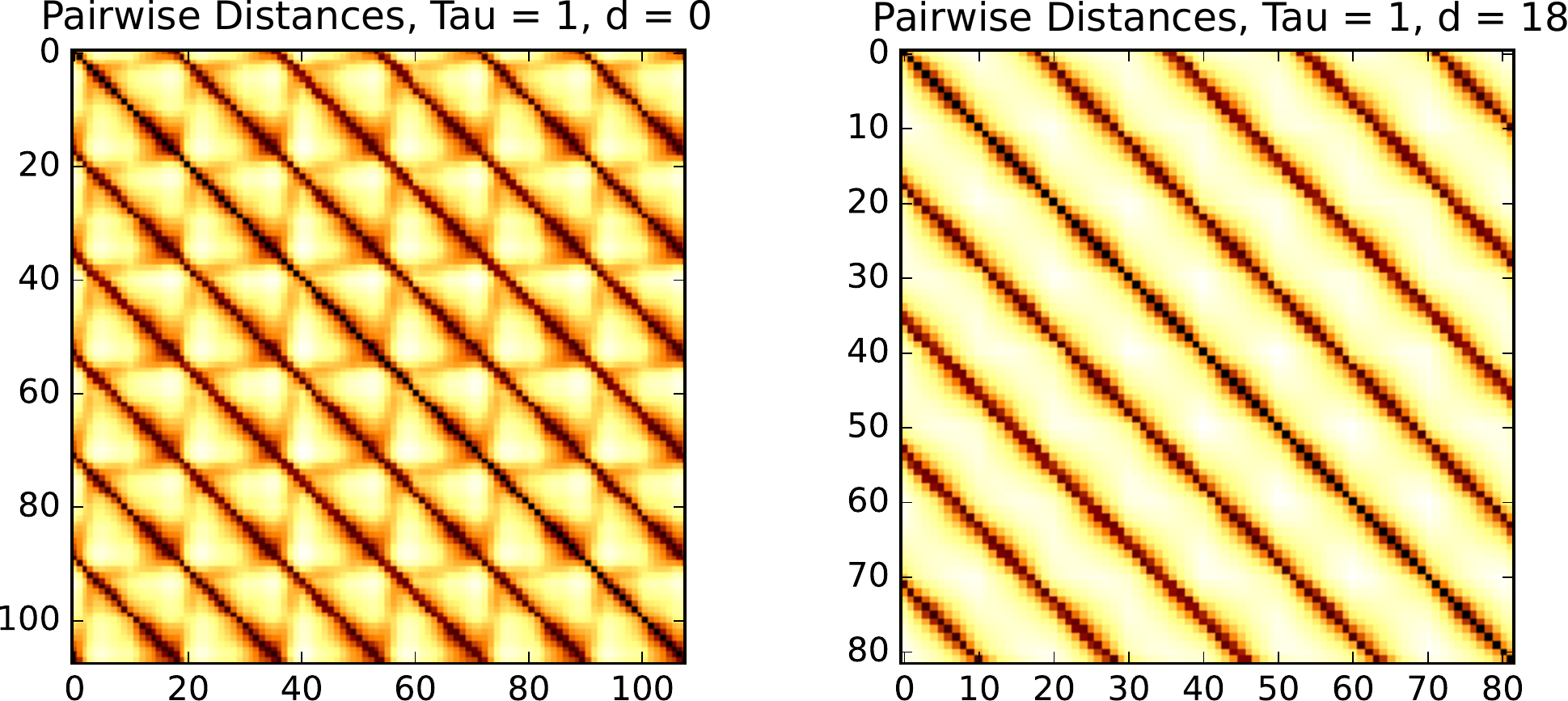}
	\caption{Self-similarity matrices $D_{Y_{0, 1}}$ and $D_{Y_{18, 1}}$ for a video of a running dog.  Even without the delay embedding ($d= 0$), the video frames still form a topological loop.  However,  a delay embedding with $d= 25$ cleans up the geometry and leads to a rounder loop, as seen in the resulting SSM.}
	\label{fig:DogRunningDelayNoDelay}
\end{figure}

Even for videos without mirror symmetries, such as a video of a running dog (Figure~\ref{fig:DogVideo}), introducing a delay brings the geometry into focus, as shown in Figure~\ref{fig:DogRunningDelayNoDelay}.

\subsection{Normalization}
\label{sec:normalization}

A few normalization steps are needed in order to enable fair comparisons between videos with different resolutions, or which have a different range in periodic motion either spatially or in intensity.  First, we perform a ``point-center and sphere normalize'' vector normalization which was shown in \cite{perea2015sliding} to have nice theoretical properties.

That is,
\begin{equation}
\widetilde{SW}_{d, \tau}(t) = \frac{ SW_{d, \tau}(t) - (SW_{d, \tau}(t)^T \boldsymbol{1}) \boldsymbol{1} }{ ||SW_{d, \tau}(t) - (SW_{d, \tau}(t)^T \boldsymbol{1}) \boldsymbol{1}||_2 }
\end{equation}
where $\boldsymbol{1}$ is a $WH(d+1) \times 1$ vector of all ones.  In other words, one subtracts the mean of each component of each vector, and  each vector is scaled so that it has unit norm (i.e. lives on the unit sphere in $\mathbb{R}^{WH(d+1)}$).  Subtracting  the mean from each component will eliminate additive linear drift on top of the periodic motion, while scaling addresses  resolution / magnitude differences.  Note that we can still use the memory optimization in Section~\ref{sec:svdmemory}, but we can no longer use the optimizations in Section~\ref{sec:diagconv} since each window is normalized independently.

Moreover, in order to mitigate nonlinear drift, we implement a simple pixel-wise convolution by the derivative of a Gaussian for each pixel in the original video before applying the delay embedding:

\begin{equation}
\hat{X}_i(t) = X_i(t) * -at \exp^{-t^2/(2\sigma^2)}
\end{equation}

This is a pixel-wise bandpass filter which could be replaced with any other bandpass filter leveraging application specific knowledge of expected frequency bounds.  This has the added advantage of reducing the number of harmonics, enabling a smaller embedding dimesion $d$.

\subsection{Periodicity/Quasiperiodicity Scoring}
\label{sec:scoring}

Once the videos are normalized to the same scale, we can score periodicity and quasiperiodicity based on the geometry of sliding window embeddings.
Let $\mathsf{dgm}_n$ be the $n$-dimensional persistence diagram
for the Rips filtration on the sliding window embedding of a video,
and define  $mp_i(\mathsf{dgm}_n)$ as the $i$-th largest difference
$d-b$ for $(b,d)\in \mathsf{dgm}_n$. In particular
\[
mp_1(\mathsf{dgm}_n) = \max\{d - b \; : \; (b,d) \in \mathsf{dgm}_n\}
\]
and $mp_i(\mathsf{dgm}_n) \geq mp_{i+1}(\mathsf{dgm}_n)$.
We propose the following scores:

\begin{enumerate}
\item {\em Periodicity Score (PS)}
\begin{equation}
\label{eq:periodicityscore}
PS = \frac{1}{\sqrt{3}} mp_1(\mathsf{dgm}_1)
\end{equation}

Like \cite{perea2015sliding}, we exploit the fact that for  the
Rips filtration on
$S^1$,  the 1-dimensional persistence
diagram has only one prominent birth-death pair with coordinates $\left(0,\sqrt{3}\right)$.
Since this is the limit shape of a normalized perfectly periodic sliding window video, the periodicity score is between 0 (not periodic) and 1 (perfectly periodic).

\item {\em Quasiperiodicity Score (QPS)}
\begin{equation}
\label{eq:quasiperiodicityscore}
QPS = \sqrt{ \frac{ mp_2(\mathsf{dgm}_1) mp_1(\mathsf{dgm}_2)  }{3} }
\end{equation}

This score is designed with the torus in mind.  We score based on the second largest 1D persistence {\em times} the largest 2D persistence, since we want a shape that has two core circles and encloses a void to get a large score.  Based on the K$\ddot{u}$nneth theorem of homology, the 2-cycle (void) should die the moment the smallest 1-cycle dies.

\item {\em Modified Periodicity Score (MPS)}
\begin{equation}
MPS = \frac{1}{\sqrt{3}} \left( mp_1(\mathsf{dgm}_1) - mp_2(\mathsf{dgm}_1) \right)
\end{equation}

We design a modified periodicity score which should be lower for quasiperiodic videos, than what the original periodicity score would yield.
\end{enumerate}

Note that we use $\mathbb{Z}_3$ field coefficients for all persistent homology computations since, as shown by \cite{perea2015sliding}, this works better for periodic signals with strong harmonics.
Before we embark on experiments, let us explore the choice of two crucial parameters for the sliding window embedding: the delay $\tau > 0$ and the dimension $d\in \mathbb{N}$.
In practice we determine an equivalent pair of parameters: the dimension $d$ and the
\emph{window size} $d\tau$.

\subsection{Dimension and Window Size}
\label{sec:windowsize}
%
%
%For simple periodic videos and as implied more generally by Taken's delay embedding theorem, a delay is necessary to for the delay embedding to be injective (free of ambiguity) in the case of 1D signals.  In video, the pixels may or may not give unique information to reconstruct the state space, so we may still need to use delay (Figure~\ref{fig:PendulumDelayNoDelay}).  But what should $M$ actually be in the worst case, if the dimensions of the state space are unknown, or in our case, if the number of independent sinusoids in the video is unknown?
Takens' embedding theorem is one of the most fundamental results
in the theory of dynamical systems \cite{takens1981detecting}.
In short, it contends that (under appropriate hypotheses) there
exists an integer $D$, so that for all $d\geq D$ and  generic $\tau >0$
the sliding window embedding $SW_{d,\tau} X$ reconstructs
the state space of the underlying dynamics witnessed by the signal
$X$.
One common strategy
for determining  a minimal such $D$ is the  \emph{false nearest-neighbors} scheme \cite{kennel1992determining}.  The idea is to keep track of the $k$-th nearest neighbors of each  point in the delay embedding, and if they change as $d$ is increased, then the prior estimates for $d$ were too low.  This algorithm was used in recent work on video dynamics \cite{venkataraman2016shape}, for instance.

\begin{figure}[!h]
	\centering
	\includegraphics[width=0.5\columnwidth]{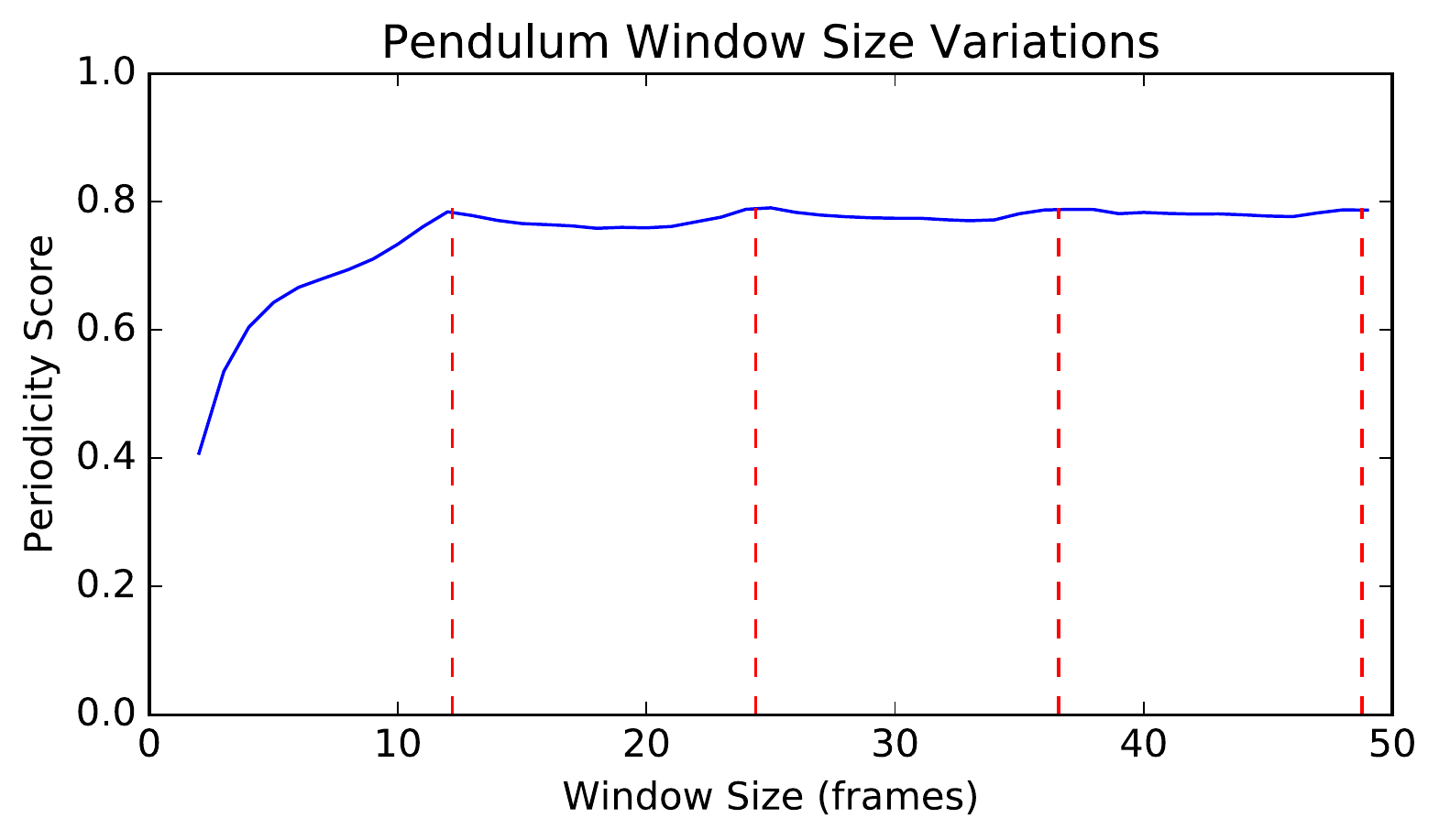}
	\caption{Varying the window size, $d\tau$, in a delay embedding of the synthetic pendulum video, which has a period length around $25$ frames.  Red dashed lines are drawn at the window lengths that would be expected to maximize roundness of the embedding for that period length based on theory in \cite{perea2015sliding}.}
	\label{fig:PendulumWindowSize}
\end{figure}

Even if we can estimate $d$, however, how does one choose the delay $\tau$?  As shown in \cite{perea2015sliding}, the sliding window embedding of a periodic signals is roundest (i.e. so that the periodicity score $PS$ is maximized) when the window size, $d \tau$, satisfies the following relation:
\begin{equation}\label{eq:windowSize}
d \tau = \frac{\pi k}{L} \left( \frac{d}{d+1} \right)
\end{equation}
Here $L$ is number of periods that  the signal has in $[0, 2\pi]$ and $k\in \mathbb{N}$.
To verify this experimentally, we show in Figure~\ref{fig:PendulumWindowSize} how the periodicity score $PS$
changes as a function of window size for the pendulum video,
and how the choice of window size from Equation \ref{eq:windowSize} maximizes
$PS$.
To generate this figure we fixed a sufficiently large $d$ and  varied $\tau$.
Let us now describe the general approach: Given a video
we perform a period-length estimation step (see section \ref{sec:autotuning} next), which results in a positive
real number $\ell$. For a given $d\in \mathbb{N}$ large enough
 we let $\tau > 0$ be so that $d\tau = \ell\cdot \frac{d}{d+1}$.

\subsection{Fundamental Frequency Estimation}
\label{sec:autotuning}

Though Figure~\ref{fig:PendulumWindowSize} suggests robustness to window size as long as the window is more than half of the period, we may not know what that is in practice.  To automate window size choices, we do a coarse estimate using fundamental frequency estimation techniques on a 1D surrogate signal.  To get a 1D signal, we extract the first coordinate of diffusion maps \cite{coifman2006diffusion} using $10 \%$ nearest neighbors on the raw video frames (no delay) after taking a smoothed time derivative.  Note that a similar diffusion-based method was also used in recent work by \cite{yaira2016no} to analyze the frequency spectrum of a video of an oscillating 2 pendulum + spring system in a quasiperiodic state.  Once we have the diffusion time series, we then apply the normalized  autocorrelation method of \cite{Mcleod05asmarter} to estimate the fundamental frequency.  In particular, given a discrete signal $x$ of length $N$, define the autocorrelation as
\begin{equation}
r_t(\tau) = \sum_{j = t}^{t + N - 1 - \tau} x_j x_{j+\tau}
\end{equation}

However, as observed by \cite{de2002yin}, a more robust function for detecting periodicities is the {\em squared difference function}
\begin{equation}
d_t(\tau) = \sum_{j = t}^{t + N - 1 - \tau} (x_j - x_{j+\tau})^2
\end{equation}
which can be rewritten as
$
d_t(\tau) = m_t(\tau) - 2r_t(\tau)
$
where
\begin{equation}
m_t(\tau) = \sum_{j = t}^{t + N - 1 - \tau} (x_j^2 + x_{j+\tau}^2)
\end{equation}

Finally, \cite{Mcleod05asmarter} suggest normalizing this function to the range $[-1, 1]$ to control for window size and to have an interpretation akin to a Pearson correlation coefficient:

\begin{equation}
\label{eq:normautocorrelation}
n_t(\tau) = 1 - \frac{m_t(\tau) - 2r_t(\tau)}{m_t(\tau)} = \frac{2r_t(\tau)}{m_t(\tau)}
\end{equation}

\begin{figure}[!htb]
	\centering \includegraphics[width=0.6\columnwidth]{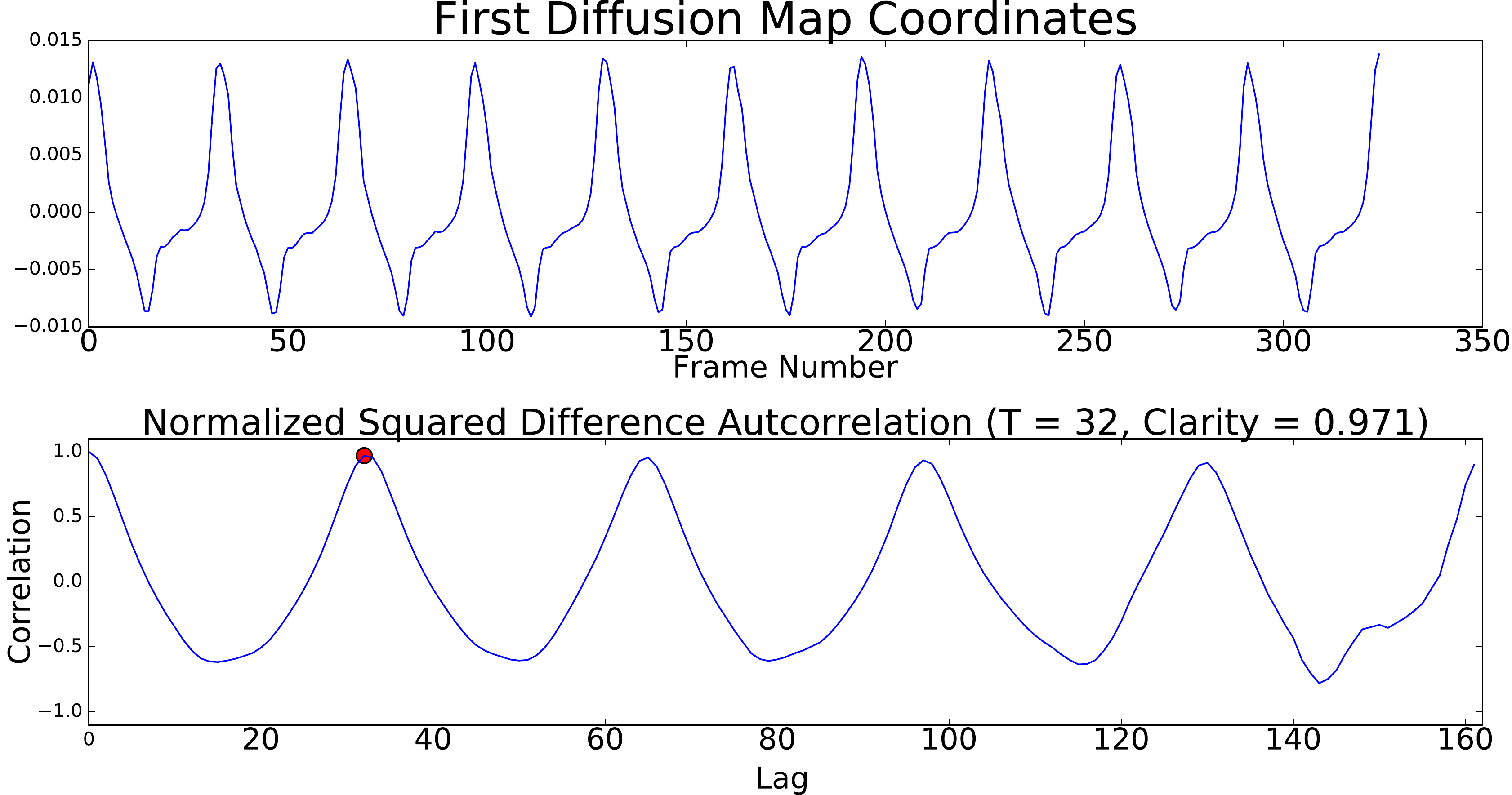}
	\caption{Diffusion maps + normalized autocorrelation fundamental frequency estimation on a periodic vocal folds video (Section~\ref{sec:vocalcords}).  The chosen period length is 32, as indicated by the red dot over the peak.  This matches with the visually inspected period length.}
	\label{fig:NormalPeriodicFundamentalFreq}
\end{figure}

\begin{figure}[!htb]
	\centering
	\includegraphics[width=0.6\columnwidth]{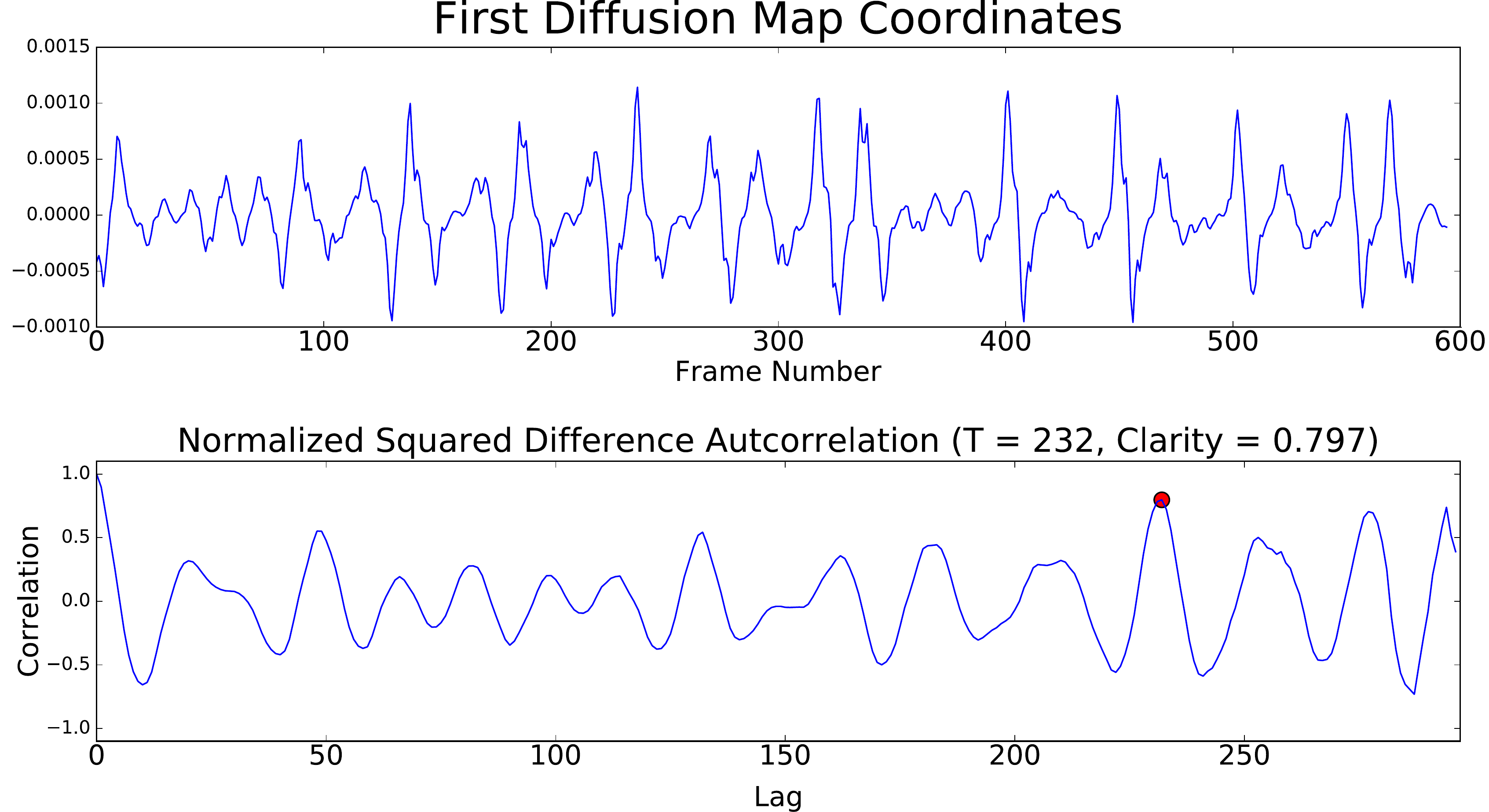}
	\caption{Diffusion maps + normalized autocorrelation fundamental frequency estimation on a video of vocal folds with irregular oscillations (Section~\ref{sec:vocalcords}).}
	\label{fig:IrregularFundamentalFreq}
\end{figure}

The fundamental frequency is then the inverse period of the largest peak in $n_t$ which is to the right of a zero crossing.  The zero crossing condition helps prevent an offset of $0$ from being the largest peak. Defining the normalized autocorrelation as in Equation~\ref{eq:normautocorrelation} has the added advantage that the value of $n_t(\tau)$ at the peak can be used to score periodicity, which the authors call \emph{clarity}. Values closer to 1 indicate more perfect periodicities.  This technique will sometimes pick integer multiples of the period, so we multiply $n_t(\tau)$ by a slowly decaying envelope which is 1 for 0 lag and 0.9 for the maximum lag to emphasize smaller periods.  Figure~\ref{fig:NormalPeriodicFundamentalFreq} shows the result of this algorithm on a periodic video,
and Figure~\ref{fig:IrregularFundamentalFreq} shows the algorithm on an irregular video.

\section{Experimental Evaluation}
Next we evaluate the effectiveness of the proposed
(Modified) Periodicity and Quasiperiodicity scores on  three different tasks.
First, we provide estimates of  accuracy  for the binary
classifications periodic/not-periodic or quasiperiodic/not-quasiperiodic in the
presence of several noise models and noise levels.
The results illustrate the robustness of our method.
Second, we quantify
the quality of periodicity rankings from machine scores, as compared to those generated by human subjects.
In a nutshell, and after comparing with several periodicity quantification algorithms,
our approach is shown to be the most closely aligned with the perception of human subjects.
Third, we demonstrate that  our methodology can be used to automatically
detect the physiological manifestations of certain speech pathologies (e.g., normal vs. biphonation),
directly from  high-speed videos of vibrating vocal folds.

\subsection{Classification Under Varying Noise Levels/Models}
\begin{figure}[!htb]
    \centering
	\subfloat[Original]{  \includegraphics[width=0.125\textwidth]{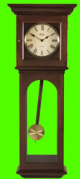} }
\quad
	\subfloat[$20\times20$ blur]{  \includegraphics[width=0.125\textwidth]{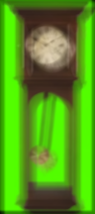} }%
\quad
	\subfloat[awgn $\sigma=2$]{  \includegraphics[width=0.125\textwidth]{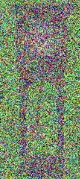} }%
\quad
	\subfloat[5\% Bit Err]{  \includegraphics[width=0.125\textwidth]{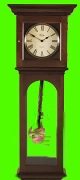} }
    \caption{The results of applying motion blur, additive white Gaussian noise, and MPEG bit corruption to a video frame.}
	\label{fig:NoiseTypes}
\end{figure}

As shown empirically in \cite{delbracio2015removing}, a common source of noise in videos comes from camera shake (blur); this is captured by point spread functions resembling  directed random walks \cite[Figure 1]{delbracio2015removing} and the amount of blur (i.e. noise level) is controlled
by the extent in pixels of the walk.
Other sources are additive white Gaussian noise (awgn), controlled by the standard deviation of the Gaussian kernel, and MPEG bit errors  quantified by the percentage of corrupted information.  Figure~\ref{fig:NoiseTypes} shows examples of these noise types.

For classification purposes we use three main recurrence classes.
Three types of periodic videos (True periodic, TP):
an oscillating pendulum, a bird flapping its wings, and an animation of a beating heart.
Two types of quasiperiodic videos (True quasiperiodic, TQ):
one showing two solid disks which oscillate sideways at non-commensurate rates, and the second showing two
stationary Gaussian pulses with amplitudes non-commensurately
modulated by cosine functions.
Two videos without significant recurrence (True non-recurrent, TN):
a video of a car driving past a landscape, and a video of an explosion.
Each one of these seven videos is then corrupted
by the three  noise models
at three different noise levels (blur = 20, 40, 80, awgn $\sigma = $ 1, 2, 3, bit error = 5, 10, 20\%) as follows:
given a particular video, a noise model and noise level,
600 instances are generated by sampling noise independently at random.
\vskip .3cm

\noindent\textbf{Results:} We report in  Table~\ref{tab:rocperiodic}
the area under the
Receiver Operating Characteristic (ROC) curve, or AUROC for short,
for the  classification task TP vs. TN (resp. TQ vs. TN)
and binary classifier furnished by  Periodicity (resp. Quasiperiodicity) Score.

\begin{table}[!htb]
  \caption{AUROC values for different levels of noise, from the binary  classification task: periodic  (bird flapping, heart beating, pendulum) vs. non-recurrent  (driving - left subcell, explosions - right subcell) based on the periodicity score (Equation~\ref{eq:periodicityscore}). Also two
synthetic quasiperiodic videos (sideways disks, modulated pulses) are compared to the same two non-recurrent videos based on the quasiperiodicity score (Equation~\ref{eq:quasiperiodicityscore}).}
  \label{tab:rocperiodic}
  \centering
  \small
  \setlength\tabcolsep{3 pt} % default value: 6pt
  \begin{tabular}{|c|c|c|c|c|c|c|c|c|c|c|c|c|c|c|c|c|c|c|}
    \hline
    &\multicolumn{2}{c|}{ }&\multicolumn{2}{c|}{ }&\multicolumn{2}{c|}{ }&\multicolumn{2}{c|}{ }&\multicolumn{2}{c|}{ }&\multicolumn{2}{c|}{ }&\multicolumn{2}{c|}{ }&\multicolumn{2}{c|}{ }&\multicolumn{2}{c|}{ } \\[-1em]
      \mbox{ }& \multicolumn{2}{c|}{\shortstack{Blur\\ 20}}& \multicolumn{2}{c|}{\shortstack{Blur \\ 40}} & \multicolumn{2}{c|}{\shortstack{Blur \\ 80}} &\multicolumn{2}{c|}{\shortstack{Awgn \\ $\sigma =1$}} & \multicolumn{2}{c|}{\shortstack{Awgn\\ $\sigma = 2$}} & \multicolumn{2}{c|}{\shortstack{Awgn \\ $\sigma =3$}} & \multicolumn{2}{c|}{\shortstack{Bit Err \\5\%}} & \multicolumn{2}{c|}{\shortstack{ Bit Err \\10\% } } & \multicolumn{2}{c|}{\shortstack{Bit Err \\ 15\%}} \\
      \hline
      \hline
      &&&&&&&&&&&&&&&&&& \\[-0.9em]
   \shortstack{Bird \\ Flapping}& 1 & 1 & 1 & 1 & 1 & 1 & 1 & 1 & 1 & 1 & 1 & 1 & 1 & 0.97 & 1 & 0.92 & 0.75 & 0.79 \\
    \hline
    &&&&&&&&&&&&&&&&&& \\[-0.9em]
   \shortstack{Heart \\ Beat} & 1 & 1 & 1 & 1 & 0.91 & 0.94 & 1 & 1 & 1 & 1 & 0.9 & 0.89 & 1 & 0.91 & 0.98 & 0.87 & 0.56 & 0.62 \\
    \hline
    &&&&&&&&&&&&&&&&&& \\[-0.6em]
   Pendulum & 1 & 0.94 & 0.84 & 0.85 & 0.43 & 0.68 & 1 & 1 & 1 & 1 & 1 & 1 & 1 & 1 & 1 & 0.99 & 0.98 & 0.97 \\
   &&&&&&&&&&&&&&&&&& \\[-0.8em]
   \hline
   \hline
   &&&&&&&&&&&&&&&&&& \\[-0.9em]
    \shortstack{QuasiPeriodic \\ Disks}& 1 & 0.85 & 1 & 0.83 & 0.75 & 0.39 & 1 & 1 & 1 & 1 & 1 & 1 & 1 & 0.92 & 0.99 & 0.93 & 0.82 & 0.82 \\
    \hline
    &&&&&&&&&&&&&&&&&& \\[-0.9em]
    \shortstack{QuasiPeriodic \\ Pulses}& 1 & 0.9 & 1 & 0.87 & 1 & 0.83 & 1 & 1 & 1 & 1 & 1 & 1 & 1 & 0.96 & 0.99 & 0.95 & 0.95 & 0.95 \\
    \hline
  \end{tabular}
  \normalsize
\end{table}

For instance, for the Blur noise model with noise level of
$80\times 80$ pixels, the AUROC from  using the
Periodicity Score to classify
the 600 instances of the Heartbeat video as periodic, and the
600 instances of the Driving video as not periodic is 0.91.
Similarly, for the MPEG bit corruption model with 5\% of bit error,
the AUROC from using the Quasiperiodicity score to classify the
600 instances of the Quasiperiodic Sideways Disks as quasiperiodic
and the 600 instances of the Explosions video as not quasiperiodic is
0.92.
To put these numbers in perspective,  AUROC = 1 is associated with a perfect classifier and  AUROC = 0.5 corresponds to  classification by a random coin flip.

Overall, the type of noise that degrades performance the most across videos is the bit error, which makes sense, since this has the effect of randomly freezing, corrupting, or even deleting frames, which all interrupt periodicity.  The blur noise also affects videos where the range of motion is small.  The pendulum video, for instance, only moves over a range of 60 pixels at the most extreme end, so an 80x80 pixel blur almost completely obscures the motion.
%Since we do not track motion, we need to be mindful of the effect of this type of motion noise present in the videos we choose to analyze.

\subsection{Comparing Human and Machine Periodicity Rankings}
\label{sec:ranking}
Next we quantify the extent to which rankings obtained from our periodicity score (Equation~\ref{eq:periodicityscore}), as well as three other methods, agree with how humans rank videos by periodicity.
The starting point is a  dataset of 20 different creative commons videos, each 5 seconds long at 30 frames per second.  Some videos appear periodic, such as a person waving hands, a beating heart, and spinning carnival rides.  Some of them appear nonperiodic, such as explosions, a traffic cam, and drone view of a boat sailing.  And some of them are in between, such as the pendulum video with simulated camera shake.

It is known that humans are notoriously bad at generating globally consistent rankings of sets with more than 5 or 7 elements \cite{miller1956magical}.
However, when it comes to binary comparisons of the type ``should A be ranked higher than B?'' few systems are as effective as human
perception, specially for the identification of recurrent patterns
in visual stimuli.  We will leverage this to generate a globally
consistent ranking of the 20 videos in our initial data set.

\begin{figure}[!htb]
    \centering
        \includegraphics[width=0.7\textwidth]{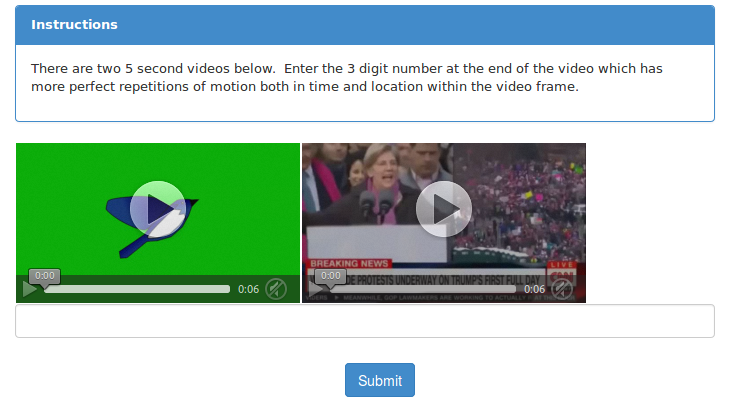}
	\caption{The interface that humans are given on AMT for pairwise ranking videos by periodicity.}
	\label{fig:turkinterface}
\end{figure}

We use Amazon's Mechanical Turk (AMT) \cite{crump2013evaluating} to
present each pair of videos in the set of 20, $\binom{20}{2} = 190$, each to three different users, for a total of 570 pairwise rankings. 15 unique AMT workers contributed to our experiment, using an interface as the one shown in Figure \ref{fig:turkinterface}.

\begin{figure}[!htb]
    \centering
    \includegraphics[width=0.6\textwidth]{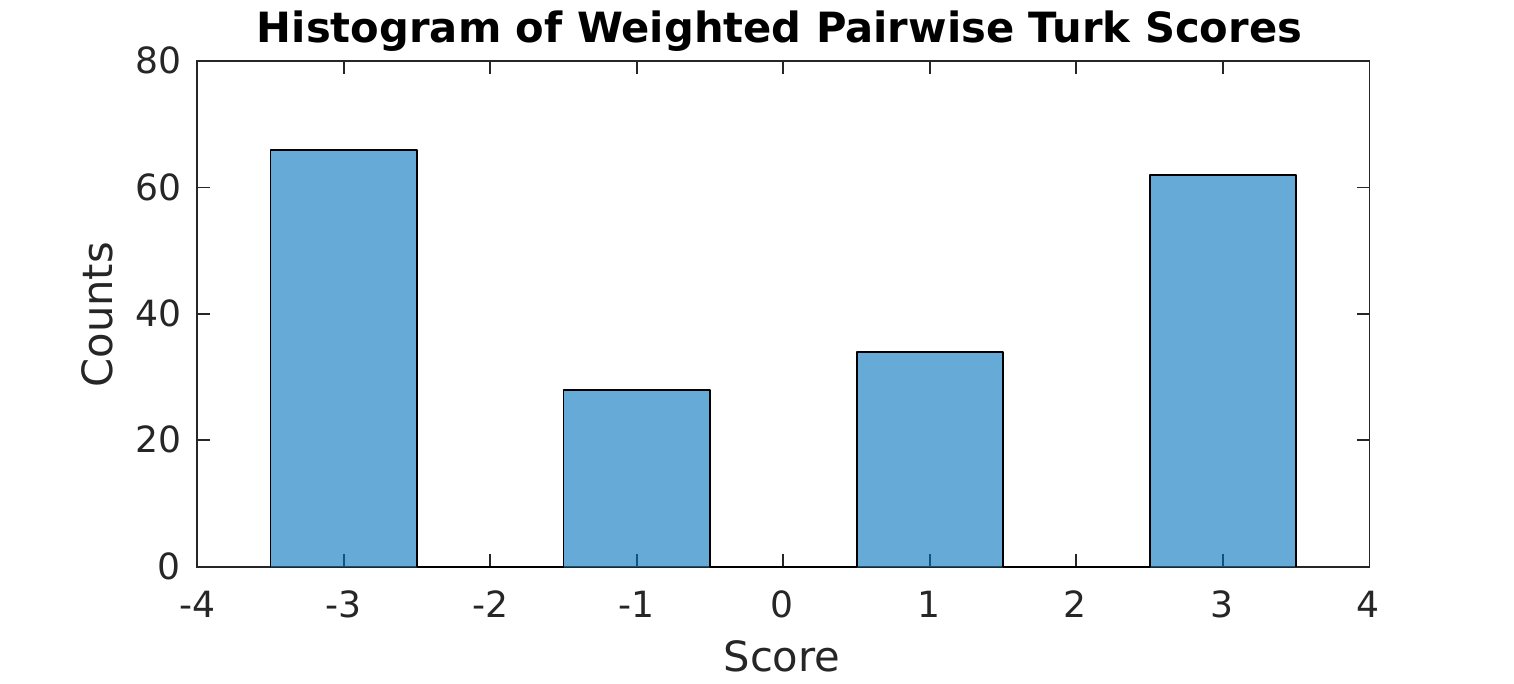}
	\caption{The histogram of scores that the workers on AMT gave to all pairwise videos.}
	\label{fig:ScoresHist}
\end{figure}

In order to aggregate this information into a global ranking which is as consistent as possible with the pairwise comparisons, we implement a technique known as {\em Hodge rank aggregation} \cite{jiang2011statistical}.
Hodge rank aggregation finds the closest consistent ranking to a set of preferences, in a least squares sense.  More precisely, given a set of objects $X$, and given a set of comparisons $P \subset X \times X$, we seek a scalar function $s$ on all of the objects that minimizes the following sum
\begin{equation}
\sum_{ (a, b) \in P} |v_{ab} - (s_b - s_a)|^2
\end{equation}
where $v_{ab}$ is a real number which is positive if $b$ is ranked higher than $a$ and negative otherwise.  Thus, $s$ is a function whose discrete gradient best matches the set of preferences  with respect to an $L^2$ norm.
Note that the preferences that we feed  the algorithm are based on the pairwise rankings returned from AMT.  If video $b$ is greater than video $a$, then we assign $v_{ab} = 1$, or -1 otherwise.  Since we have 3 rankings for each video, we actually assign weights of +3, +1, -1, or -3.  The +/- 3 are if all rankings agree in one direction, and the +/- 1 are if one of the rankings disagrees with the other two.  Figure~\ref{fig:ScoresHist} shows a histogram of all of the weighted scores from users on AMT.  They are mostly in agreement, though there are a few +/- 1 scores.

As comparison to the human scores, we use three different classes of techniques for machine ranking of periodicity.
\vskip .3cm

\noindent\emph{Sliding Windows (SW):}
We sort the videos in decreasing order of Periodicity Score (Equation~\ref{eq:periodicityscore}).  We fix the window size at 20 frames and the embedding dimension at 20 frames (which is enough to capture 10 strong harmonics).  We also apply a time derivative of width 10 to every frame.
\vskip .3cm

\noindent\emph{Cutler-Davis \cite{cutler2000robust}:}
The authors of this work present two different techniques to quantify periodicity from a self-similarity matrix (SSM) of video frames.  The first is a frequency domain technique based on the peak of the average power spectral density over all columns (rows) of the SSM after linearly de-trending and applying a Hann window.  To turn this into a continuous score, we  report the ratio of the peak minus the mean over the standard deviation.
This method will be referred to as \emph{Frequency Score}.

As the authors warn, the frequency peak method has a high susceptibility to false positives.  This motivated the design of a more robust technique in \cite{cutler2000robust}, which works by finding peaks in the 2D normalized autocorrelation of the Gaussian smoothed SSMs.  For videos with mirror symmetry, the peaks will lie on a diamond lattice, while for videos without mirror symmetry, they will lie on a square lattice.
After peak finding within neighborhoods, one simply searches over all possible lattices at all possible widths to find the best match with the peaks.  Since each lattice is centered at the autocorrelation point $(0, 0)$, no translational checks are necessary.

%\begin{figure}[!htb]
%	\centering
%	\includegraphics[width=\columnwidth]{PendulumCutlerDavis.pdf}
%	\caption{The method of \cite{cutler2000robust} on the pendulum video.  The peak PSD frequency is above two times the standard deviation above the mean (green line, center figure).  The maxes (green points) on the 2D autocorrelation of the SSM lie on a diamond shape pattern.}
%	\label{fig:PendulumCutlerDavis}
%\end{figure}

%\begin{figure}[!htb]
%	\centering	\includegraphics[width=\columnwidth]{DogRunningCutlerDavis.pdf}
%	\caption{The method of \cite{cutler2000robust} on the video of a running dog.  The peak PSD frequency is above two times the standard deviation above the mean.  The maxes (green points) of the 2D autocorrelation of the SSM lie on a square pattern, since there is no symmetry.  However, there are also spurious points along the diagonal (orange points), which can happen due to numerical instability for finding maxes along diagonal lines with a near constant max value.}
%	\label{fig:DogRunningCutlerDavis}
%\end{figure}

To turn this into a continuous score, let $E$ be the sum of  Euclidean distances of the matched peaks in the autocorrelation image to the best fit lattice, let $r_1$ be the proportion of lattice points that have been matched, and let $r_2$ be the proportion of peaks which have been matched to a lattice point.  Then we give the final periodicity score as
\begin{equation}
CD_{score} = (1 + E/r_1)/(r_1 r_2)^3
\end{equation}

A lattice which fits the peaks perfectly ($r_1 = 1$) with no error ($E = 0$) and no false positive peaks ($r_2 = 1$) will have a score of 1, and any video which fails to have a perfectly matched lattice will have a score greater than 1.  Hence, we sort in increasing order of the score to get a ranking.

As we will show, this technique agrees the second best with humans after our Periodicity Score ranking.  One of the main drawbacks is numerical stability of finding maxes in non-isolated critical points around nearly diagonal regions in square lattices, which will erroneously inflate the score.  Also, the lattice searching only occurs over an integer grid, but there may be periods that aren't integer number of frames, so there will always be a nonzero $E$ for such videos.  By contrast, our sliding window scheme can work for any real valued period length.
\vskip .3cm

\noindent\emph{Diffusion Maps + Normalized Autocorrelation ``Clarity'':}
Finally, we apply the technique from Section~\ref{sec:autotuning} to get an autocorrelation function, and we report the value of the maximum peak of the normalized autocorrelation to the right of a zero crossing, referred to as ``clarity'' by \cite{Mcleod05asmarter}.  Values closer to 1 indicate more perfect repetitions, so we sort in descending order of clarity to get a ranking.
\vskip .3cm

\begin{figure}[!htb]
    \centering
    \subfloat{
        \includegraphics[width=0.6\textwidth]{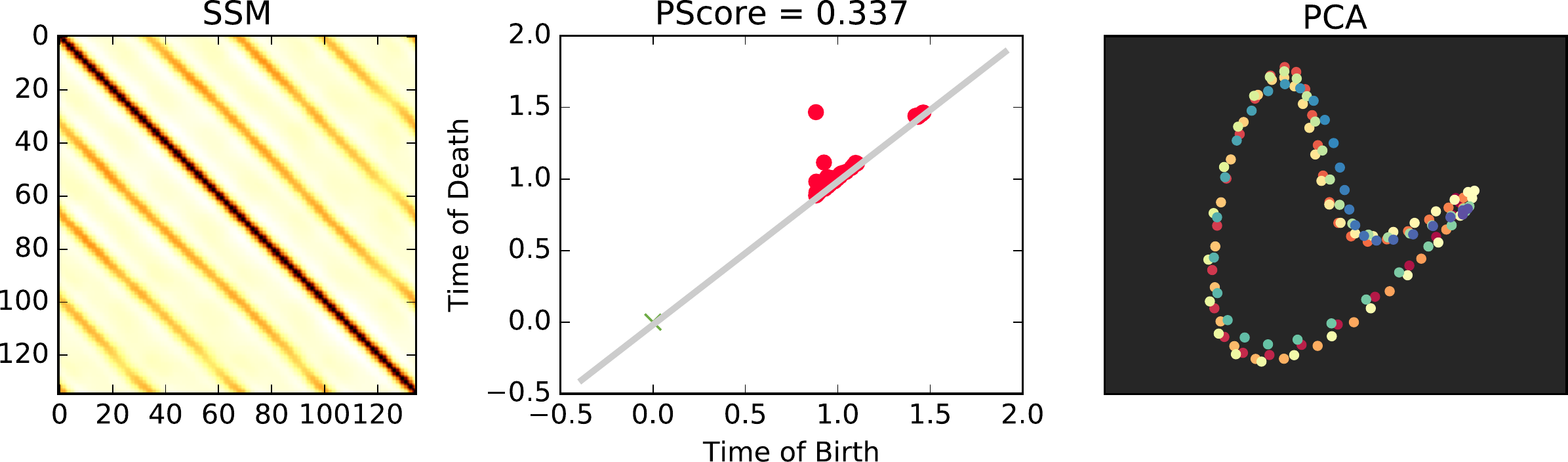}
        }\\
    \subfloat{
        \includegraphics[width=0.3\textwidth]{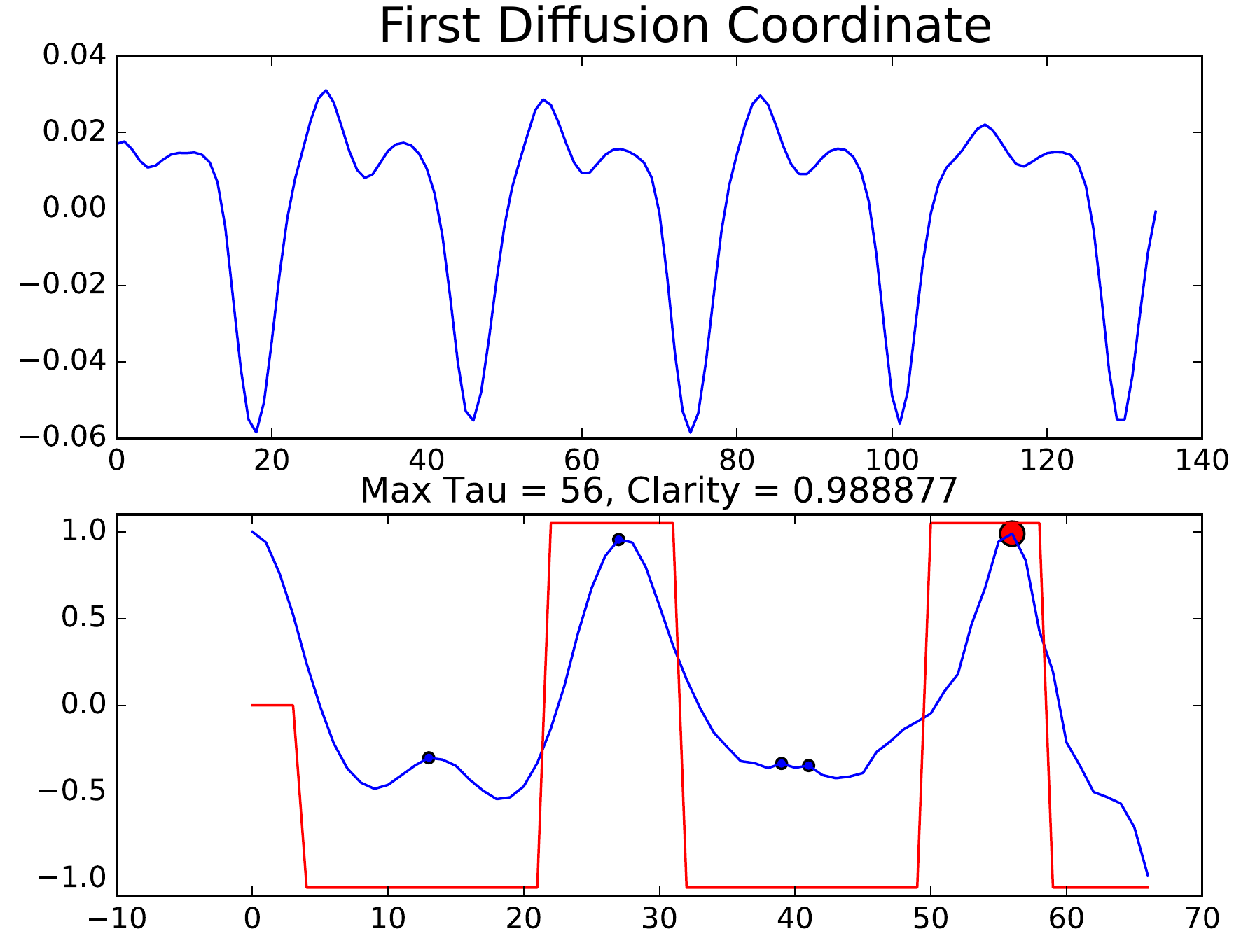}
        }
	\subfloat{
        \includegraphics[width=0.5\textwidth]{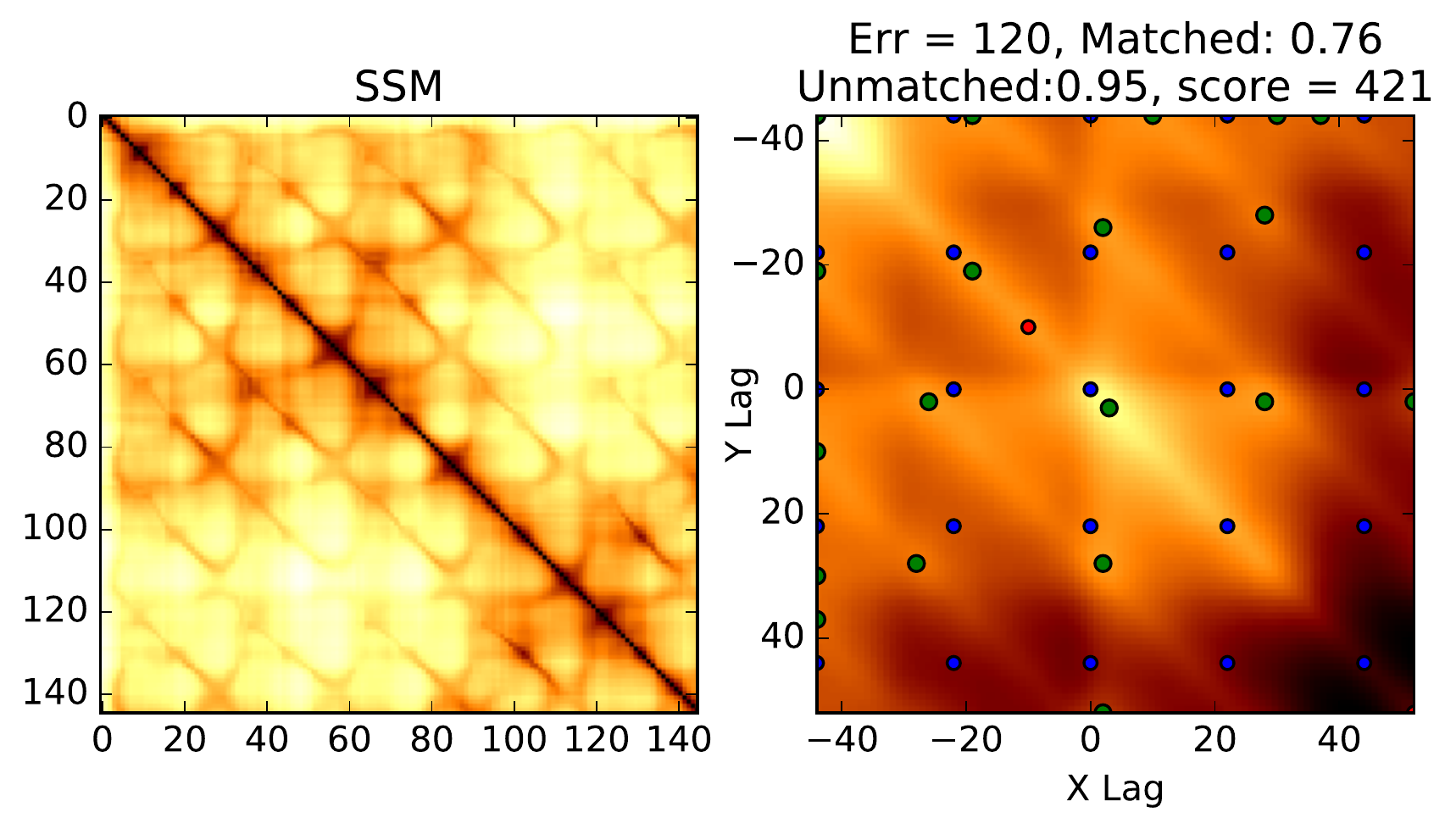}
        }
	\caption{An example of the SW score (top), the clarity score (bottom left), and the $CD_{score}$ (bottom right, matched peaks in green and lattice in blue), on a periodic video of a man waving his arms from the KTH dataset (\cite{schuldt2004recognizing}).}
	\label{fig:PeriodicScoresExample}
\end{figure}

Figure~\ref{fig:PeriodicScoresExample} shows an example of these three different techniques on a periodic video.  There is a dot which rises above the diagonal in the persistence diagram, a lattice is found which nearly matches the critical points in the autocorrelation image, and autocorrelation function on diffusion maps has a nice peak.

\begin{figure}[!htb]
    \centering
    \subfloat{
        \includegraphics[width=0.6\textwidth]{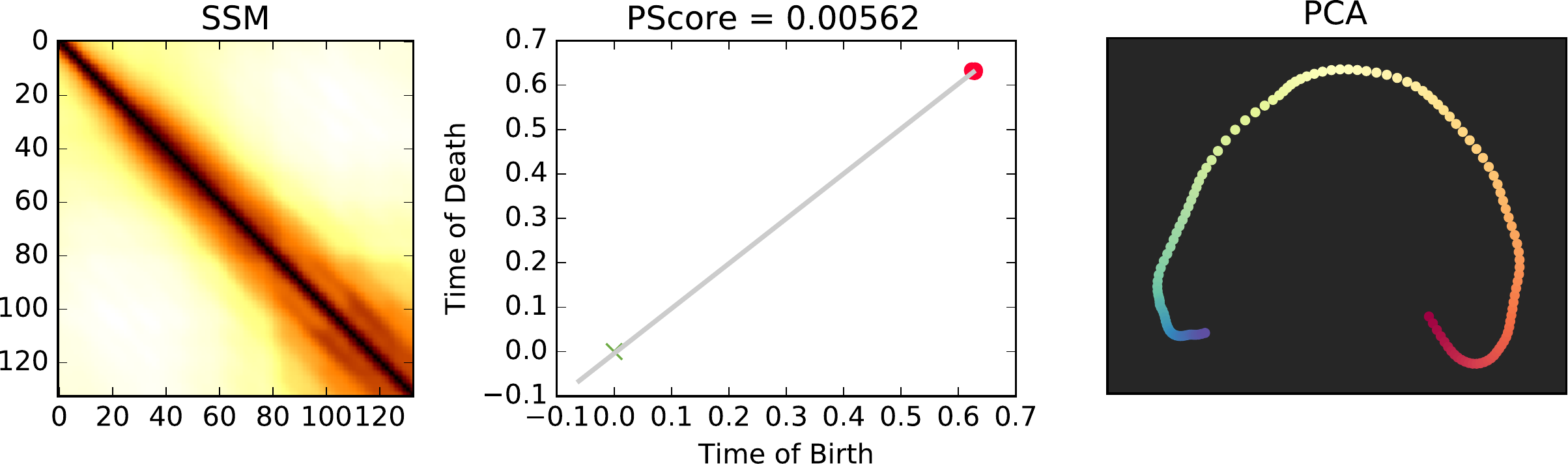}
        }\\
    \subfloat{
        \includegraphics[width=0.3\textwidth]{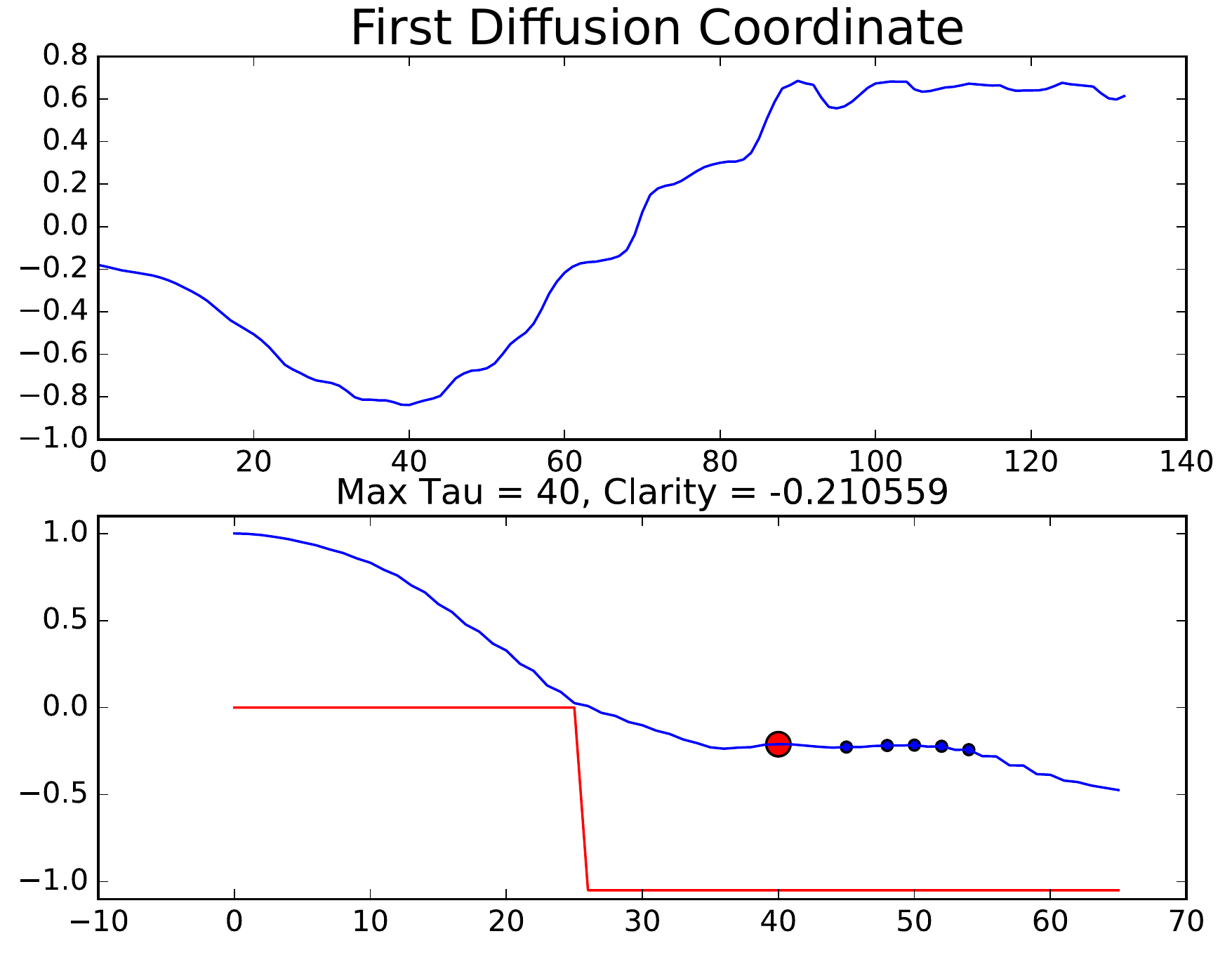}
        }
	\subfloat{
        \includegraphics[width=0.5\textwidth]{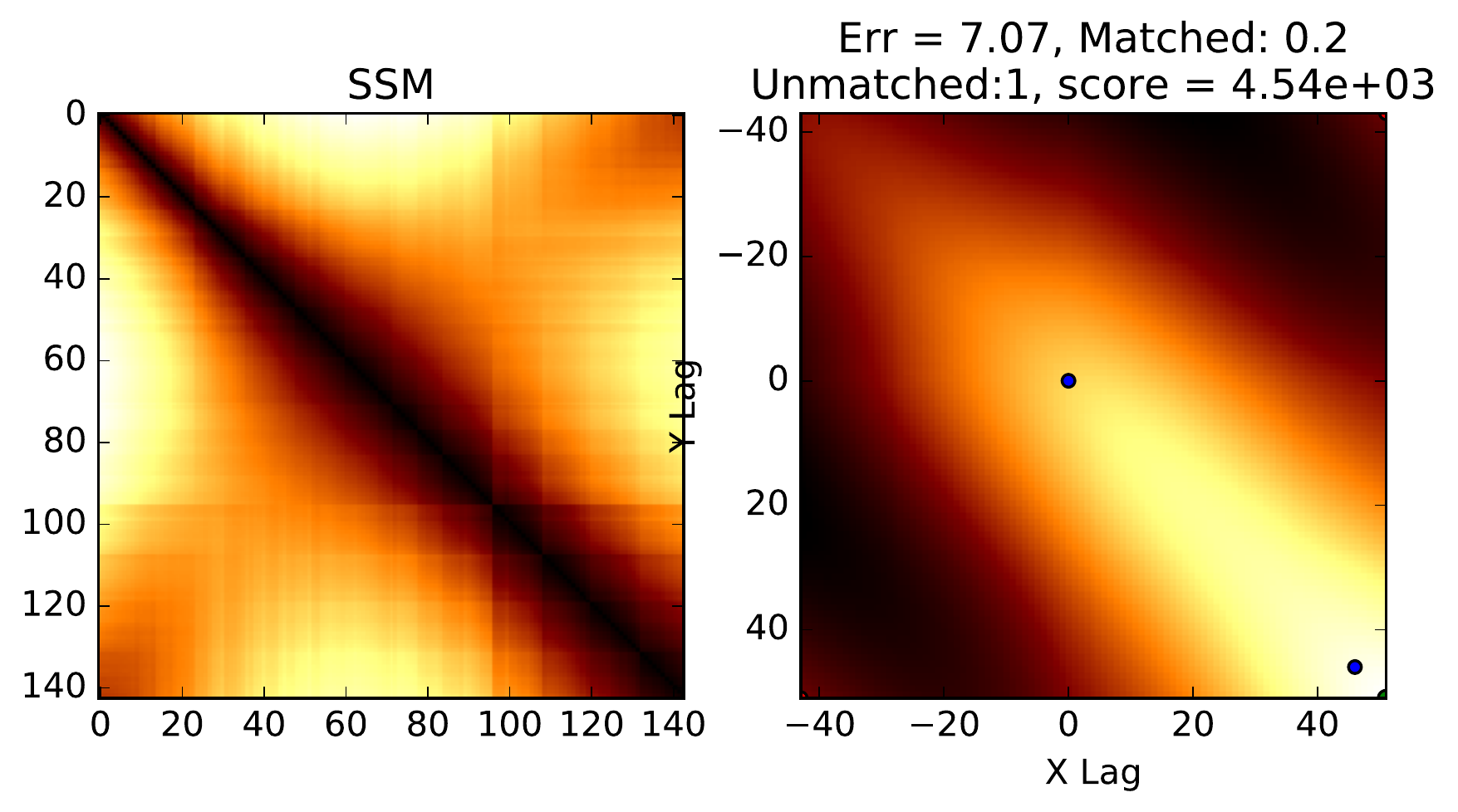}
        }
	\caption{An example of the SW score (top), the clarity score (bottom left), and the $CD_{score}$  (bottom left, matched peaks in green and lattice in blue), on a video of an explosion, which is nonperiodic.}
	\label{fig:NonPeriodicScoresExample}
\end{figure}

By contrast, for a nonperiodic video (Figure~\ref{fig:NonPeriodicScoresExample}), there is hardly any persistent  homology, there is no well matching lattice, and the first diffusion coordinate has no apparent periodicities.
\vskip .3cm

\noindent\textbf{Results:}
Once we have the global human rankings and the global machine rankings, we can compare them using the {\em Kendall $\tau$ score} \cite{kendall1938new}.  Given a set of objects $N$ objects $X$ and two total orders $>_1$ and $>_2$, where $>(x_a, x_b) = 1$ if $x_a > x_b$ and $>(x_a, x_b) = -1$ if $x_a < x_b$, the Kendall $\tau$ score is defined as

\begin{equation}
\tau = \frac{1}{N(N-1)/2} \sum_{i < j} (>_1(x_i, x_j)) (>_2(x_i, x_j))
\end{equation}

For two rankings which agree exactly, the Kendall $\tau$ score will be 1.  For two rankings which are exactly the reverse of each other, the Kendall $\tau$ score will be -1.  In this way, it analogous to a Pearson correlation between rankings.

\begin{table}[htbp]
\centering
\caption{The Kendall $\tau$ scores between all of the machine rankings and the Hodge aggregated human rankings.}
\begin{tabular}{|l|r|r|r|r|r|r|}
\hline
\multicolumn{1}{|r|}{} & \multicolumn{1}{l|}{Human} & \multicolumn{1}{l|}{SW+TDA} & \multicolumn{1}{l|}{Freq\cite{cutler2000robust}} & \multicolumn{1}{l|}{CD$_{score}$\cite{cutler2000robust}} & \multicolumn{1}{l|}{Clarity\cite{de2002yin}} \\ \hline
Human & 1 & \textbf{0.663} & -0.295 & 0.347 & 0.284 \\ \hline
%TDA $Z_2$ & 0.642 & 1 & 0.979 & -0.295 & 0.2 & 0.537 \\ \hline
SW+TDA  & 0.663 & 1 & -0.316 & 0.221 & \textbf{0.516} \\ \hline
Freq\cite{cutler2000robust} & -0.295 & -0.316 & 1 & -0.0842 & -0.189 \\ \hline
CD$_{score}$\cite{cutler2000robust} & 0.347 & 0.221 & -0.0842 & 1 & 0.411 \\ \hline
Clarity\cite{de2002yin} & 0.284 & 0.516 & -0.189 & 0.411 & 1 \\ \hline
\end{tabular}
\label{tab:globalrankcomparisons}
\end{table}

\begin{table}[htbp]
\centering
\caption{Average runtimes, in milliseconds, per video for all of the algorithms}
\begin{tabular}{|r|r|r|r|r|r|}
\hline
 \multicolumn{1}{|l|}{SW+TDA} & \multicolumn{1}{l|}{Freq\cite{cutler2000robust}} & \multicolumn{1}{l|}{CD$_{score}$\cite{cutler2000robust}} & \multicolumn{1}{l|}{Clarity\cite{de2002yin}} \\ \hline
3101ms & 73ms & 176ms & 154ms \\ \hline
\end{tabular}
\label{tab:videoranktimes}
\end{table}

Table~\ref{tab:globalrankcomparisons} shows the Kendall $\tau$ scores between all of the different machine rankings and the human rankings. Our sliding window video methodology (SW+TDA) agrees with the human ranking more than any other pair of ranking types.  The second most similar are the SW and the diffusion clarity, which is noteworthy as they are both geometric techniques.  Table~\ref{tab:videoranktimes} also shows the average run times, in milliseconds, of the different algorithms on each video on our machine.  This does highlight one potential drawback of our technique, since TDA algorithms tend to be computationally intensive.  However, at this scale (videos with at most several hundred frames), performance is reasonable.

\subsection{Periodicity And Biphonation in High Speed Videos of Vocal Folds}
\label{sec:vocalcords}
In this final task we apply our methodology to a real world problem of interest in medicine.  We show that our method can automatically detect certain types of voice pathologies from high-speed glottography, or high speed videos (4000 fps) of the left and right vocal folds in the human vocal tract \cite{wittenberg1995recording,wilden1998subharmonics,deliyski2007clinical}.  In particular, we detect and differentiate quasiperiodicity from periodicity by using our geometric sliding window pipeline.  Quasiperiodicity is a special case of what is referred to as ``biphonation'' in the biological context, where nonlinear phenomena cause a physical process to bifurcate into two different periodic modes, often during a transition to chaotic behavior \cite{herzel1996biphonation}.  The torus structure we sketched in Figure~\ref{fig:Quasiperiodic1D} has long been recognized in this context \cite{neubauer2001spatio}, but we provide a novel way of quantifying it.

Similar phenomena exist in audio \cite{herzel1994analysis, herzel1996biphonation}, but the main reason for studying laryngeal high speed video is understanding the biomechanical underpinnings of what is perceived in the voice. In particular, this understanding can potentially lead to practical corrective therapies and surgical interventions. On the other hand, the presence of biphonation in sound is not necessarily the result of a physiological phenomenon; it has been argued that it may come about as the result of changes in states of arousal \cite{briefer2015segregation}.

In contrast with our work, the existing literature on video-based techniques usually employs an inherently {\em Lagrangian} approach, where different points on the left and right vocal folds are tracked, and coordinates of these points are analyzed as 1D time series (e.g. \cite{neubauer2001spatio, qiu2003automatic, lohscheller2007clinically, herbst2016phasegram}, \cite{mehta2011automated}).  This is a natural approach, since those are the pixels where all of the important signal resides, and well-understood 1D signal processing technique can be used.  However, edge detectors often require tuning, and they can suddenly fail when the vocal folds close \cite{lohscheller2007clinically}.  In our technique, we give up the ability to localize the anomalies (left/right, anterior/posterior) since we are not tracking them, but in return we do virtually no preprocessing, and our technique is domain independent.
%
%There are also differences in how we quantify the signals.  \cite{neubauer2001spatio} perform PCA on all of the time series from the tracked 1D points on the vocal folds, and they examine the Fourier spectrum of the principal components to qualitatively point out differences in the modes present between normal phonation and biphonation.  Many works also use standard techniques from nonlinear time series analysis on these signals, such as phase portraits, Poincar\'{e} Sections, and next amplitude maps \cite{herzel1994analysis}.  A recent work uses the entropy of histograms of 1D Poincare Sections to detect bifurcations and the onset of irregularities \cite{herbst2016phasegram}.  However, even though bifurcations can be effectively detected, they only use a 1-lag time delay embedding, which is inadequate to properly reconstruct the quasiperiodic phase space.  In our work, we specifically reconstruct and detect quasiperiodic phase spaces with a joint multi-lag delay embedding of all pixels using tools which can work in high dimensional ambient spaces.  This allows us, for example, to differentiate between Figure~\ref{fig:VocalCordsBiphonation} and Figure~\ref{fig:VocalCordsIrregular}.
\vskip .3cm

\begin{figure}[!htb]
    \centering
    \subfloat{
        \includegraphics[width=0.8\columnwidth]{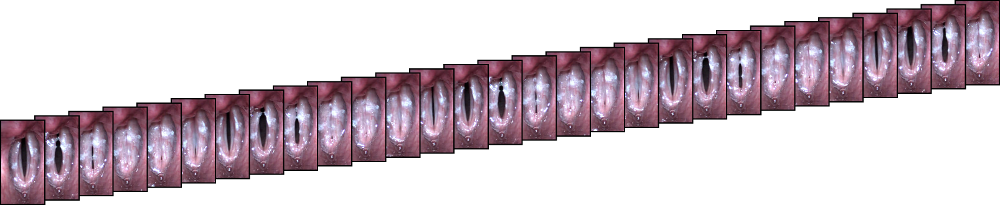}
        }\\
    \subfloat{
        \includegraphics[width=0.8\columnwidth]{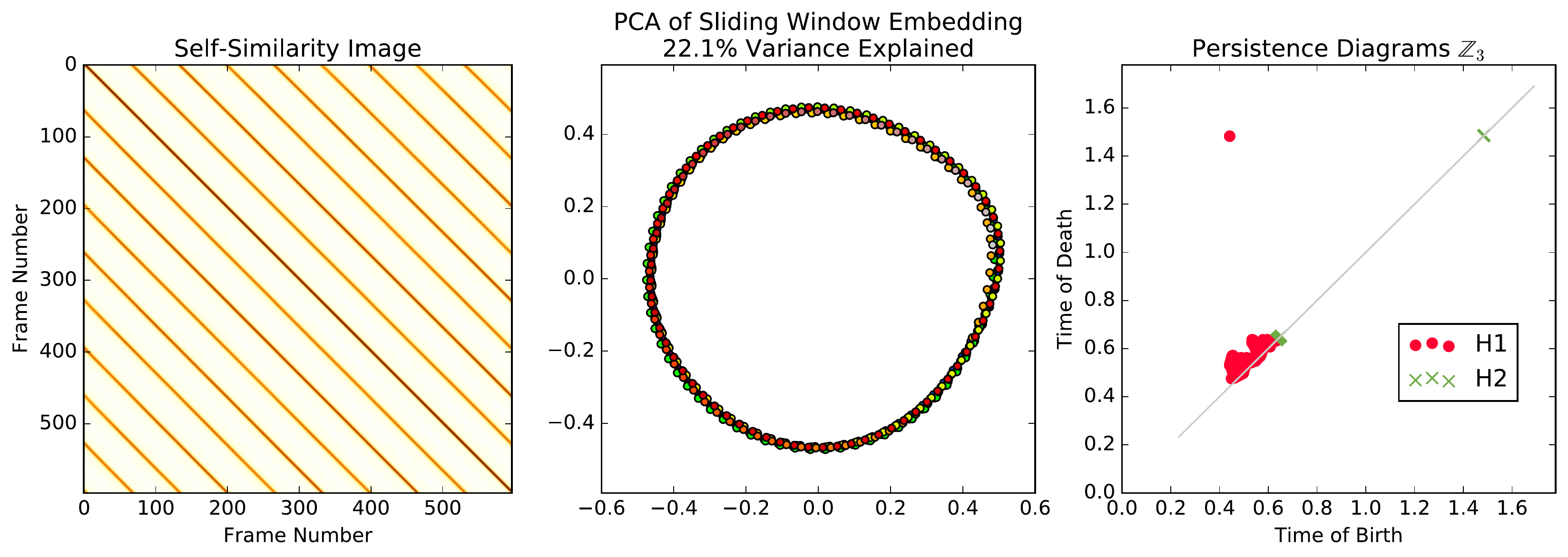}
        }
	\caption{Video frames and sliding window statistics on a video of vocal folds undergoing normal periodic vibrations \cite{mehta2011automated}.  One strong loop is visible in PCA and in the persistence diagrams}
	\label{fig:VocalCordsNormalPeriodic}
\end{figure}

\begin{figure}[!htb]
    \centering
    \subfloat{
        \includegraphics[width=0.8\columnwidth]{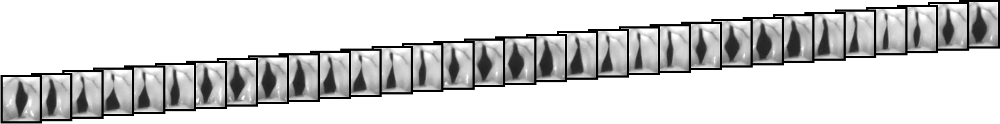}
        }\\
    \subfloat{
        \includegraphics[width=0.8\columnwidth]{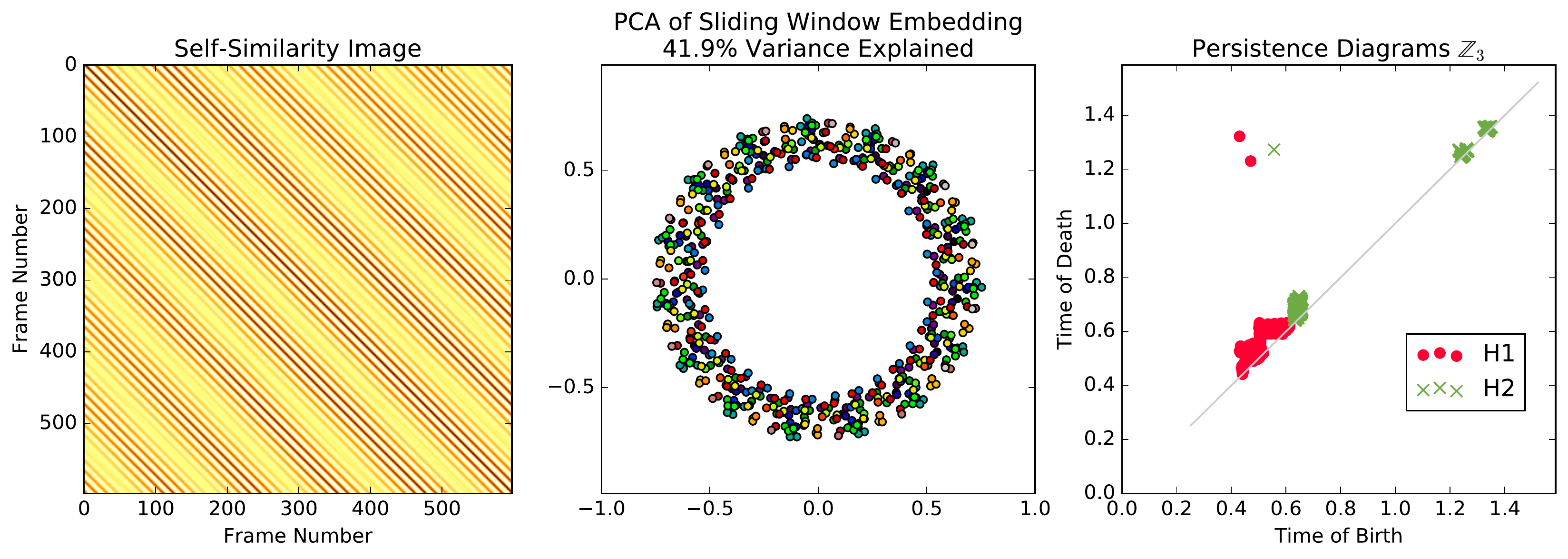}
        }
	\caption{Video frames and sliding window statistics on a video of vocal folds undergoing biphonation, courtesy of Juergen Neubauer \cite{neubauer2001spatio}.  PCA suggests a possible torus, and the persistence diagram indeed has the signature of a torus (two strong independent 1-cycles and one 2-cycle)}
	\label{fig:VocalCordsBiphonation}
\end{figure}

\begin{figure}[!htb]
    \centering
    \subfloat{
        \includegraphics[width=0.8\columnwidth]{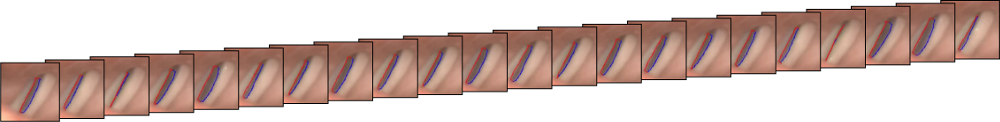}
        }\\
    \subfloat{
        \includegraphics[width=0.8\columnwidth]{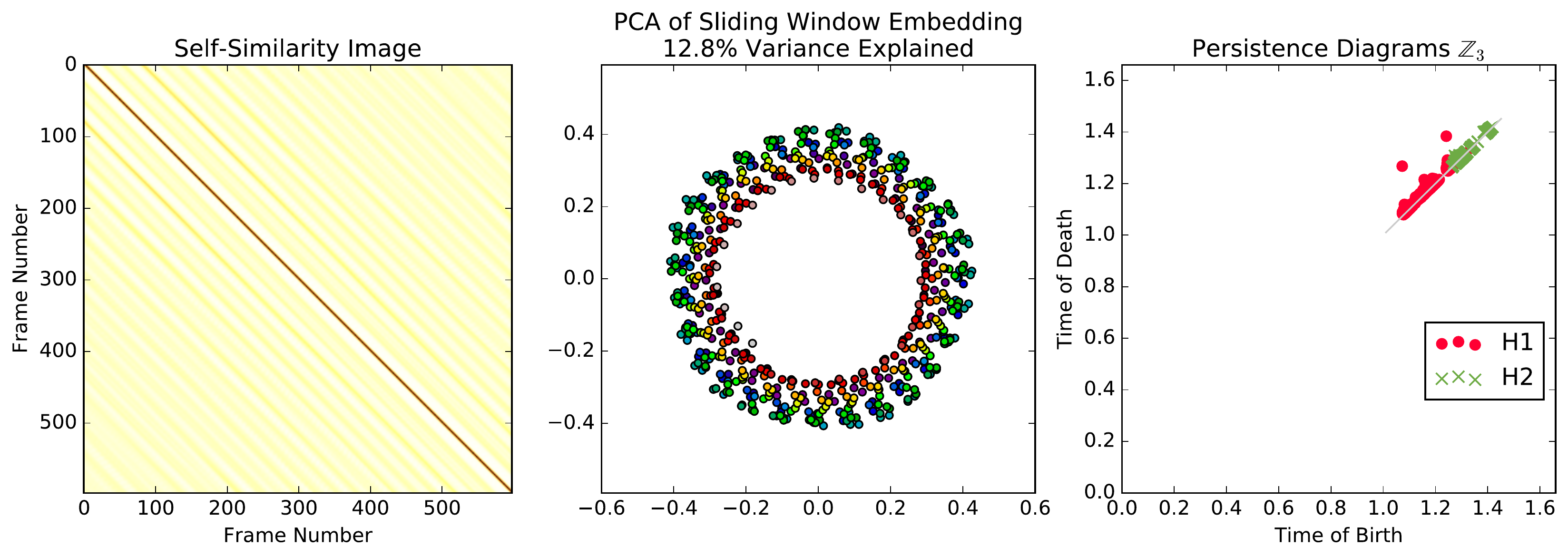}
        }
	\caption{Video frames and sliding window statistics of ``irregular'' vocal fold vibrations \cite{herbst2016phasegram}.  Though 2D PCA looks similar to Figure~\ref{fig:VocalCordsBiphonation}, no apparent 1D or 2D topological features are apparent in the high dimensional state space.}
	\label{fig:VocalCordsIrregular}
\end{figure}

\noindent\textbf{Results:} We use a collection of 7 high-speed videos for this analysis, drawn from a variety of different sources \cite{zacharias2016comparison}, \cite{mehta2011automated}, \cite{neubauer2001spatio}, \cite{herbst2016phasegram}.  There are two videos which correspond to ``normal'' periodic vocal folds, three which correspond to biphonation \cite{neubauer2001spatio}, and two which correspond to irregular motion\footnote{Please refer to supplementary material for an example video from each of these three classes}.  We manually extracted 400 frames per video (100milliseconds) and autotuned the window size based on autocorrelation of 1D diffusion maps (Section~\ref{sec:autotuning}).  We then chose an appropriate $\tau$  and  chose a time spacing so that each point cloud would have 600 points.  As shown in Table~\ref{tab:VocalCordResults}, our technique is able to differentiate between the four classes.  We also show PCA and persistence diagrams for one example for each class.  In Figure~\ref{fig:VocalCordsNormalPeriodic}, we see what appears to be a loop in PCA, and one strong 1D persistent dot confirms this.  In Figure~\ref{fig:VocalCordsBiphonation}, we see a prominent torus in the persistence diagram.  In Figure~\ref{fig:VocalCordsIrregular}, we don't see any prominent structures in the persistence diagram, even though PCA looks like it could be a loop or a torus.  Note, however, that PCA only preserves 13.7\% of the variance in the signal, which is why high dimensional techniques are important to draw quantitative conclusions.
\begin{table}[!htbp]
\centering
\caption{Results of our sliding window pipeline on videos of periodic vocal folds, biphonation, and irregularities.  We give the max persistence periodicity score (PS), the modified periodicity score (MPS), the harmonic score (HS), and quasiperiodic score (QPS) presented in Section~\ref{sec:scoring}.  We also show the window size (Win) that the autocorrelation technique in Section~\ref{sec:autotuning} gives.  We have bolded the top three MPS and QPS scores across all videos.  The max modified periodic scores include the two periodic videos and one of the biphonation videos.  The max quasiperiodic scores are all of the biphonation videos, which means the one with a high periodicity score could be ruled out of the periodicity category.}
\begin{tabular}{|l|c|c|c|c|}
\hline
Video Name & \multicolumn{1}{l|}{Win} & \multicolumn{1}{l|}{PS} & \multicolumn{1}{l|}{MPS} & \multicolumn{1}{l|}{QPS} \\ \hline
Periodic 1 \scriptsize (\cite{herbst2016phasegram}) & 16 & 0.816 & \textbf{0.789} & 0.011 \\ \hline
Periodic 2 \scriptsize (\cite{mehta2011automated}, Figure~\ref{fig:VocalCordsNormalPeriodic}) & 32 & 0.601 & \textbf{0.533} & 0.009 \\ \hline
Biphonation 1 \scriptsize (\cite{neubauer2001spatio}) & 53 & 0.638 & 0.294 & \textbf{0.292} \\ \hline
Biphonation 2 \scriptsize (\cite{neubauer2001spatio}) & 42 & 0.703 & \textbf{0.583} & \textbf{0.116} \\ \hline
Biphonation 3 \scriptsize (\cite{neubauer2001spatio}, Figure~\ref{fig:VocalCordsBiphonation}) & 67 & 0.515 & 0.076 & \textbf{0.426} \\ \hline
Mucus Perturbed Periodic \scriptsize (\cite{zacharias2016comparison}) & 94 & 0.028 & 0.019 & 0.004 \\ \hline
Irregular \scriptsize (\cite{herbst2016phasegram}, Figure~\ref{fig:VocalCordsIrregular}) & 232 & 0.18 & 0.097 &  0.04 \\ \hline
\end{tabular}
\label{tab:VocalCordResults}
\end{table}

\section{Discussion}
We have shown in this work how  applying sliding window embeddings
to videos can be used to
translate properties of the underlying dynamics into
geometric features of the resulting point cloud representation.
Moreover, we also showed how topological/geometric tools
such as persistence homology can be leveraged to
quantify the geometry of these embeddings.
The pipeline was evaluated extensively showing robustness to several noise models, high quality in the produced
periodicity rankings and applicability to the
study of speech conditions form high-speed video data.

%Our attempt to correct for drift can be extended to tackle more complicated camera shake / alignment scenarios.  The challenge in that case will be to blindly decouple drift from legitimate periodic motion.  We remark that other Eulerian approaches to analyzing periodic videos suffer from similar problems when drift is present \cite{wadhwa2013phase}.  If enough of the signal is in color changes in the video as opposed to motion, a simple solution is to use optical flow to estimate motion and to warp all frames to a canonical pose with a piecewise affine map.  This has been done, for instance, in automatic heart-rate estimation algorithms in consumer video \cite{kumar2015distanceppg, tulyakov2016self}.

Moving forward, an interesting avenue related to medical applications is the difference between biphonation which occurs from quasiperiodic modes and biphonation which occurs from harmonic modes.  \cite{perea2015sliding} shows that $\mathbb{Z}_3$ field coefficients can be used to indicate the presence of a strong harmonic, so we believe a geometric approach is possible.  This could be used, for example, to differentiate between subharmonic anomalies and quasiperiodic transitions \cite{wilden1998subharmonics}.
%We would then like to combine our techniques with those in \cite{herbst2016phasegram} on much larger databases to see if a large number of different types of anomalies can be automatically detected and classified.

\section*{Acknowledgments}
The authors would like to thank Juergen Neubauer,  Dimitar Deliyski, Robert Hillman, Alessandro de Alarcon, Dariush Mehta, and Stephanie Zacharias for providing videos of vocal folds.  We also thank Matt Berger at ARFL for discussions about sliding window video efficiency, and we thank the 15 anonymous workers on the Amazon Mechanical Turk who ranked periodic videos.

\bibliographystyle{plain}
\bibliography{main}

\begin{thebibliography}{10}

\bibitem{allmen1990cyclic}
Mark Allmen and Charles~R Dyer.
\newblock Cyclic motion detection using spatiotemporal surfaces and curves.
\newblock In {\em Pattern Recognition, 1990. Proceedings., 10th International
  Conference on}, volume~1, pages 365--370. IEEE, 1990.

\bibitem{atanbori2013analysis}
John Atanbori, Peter Cowling, John Murray, Belinda Colston, Paul Eady, Dave
  Hughes, Ian Nixon, and Patrick Dickinson.
\newblock Analysis of bat wing beat frequency using fourier transform.
\newblock In {\em International Conference on Computer Analysis of Images and
  Patterns}, pages 370--377. Springer, 2013.

\bibitem{Ripser}
Ulrich Bauer.
\newblock {Ripser}: a lean {C++} code for the computation of {Vietoris–Rips}
  persistence barcodes.
\newblock \url{http://ripser.org}, 2015–2017.

\bibitem{coifman2006diffusion}
Ronald~R Coifman and St{\'e}phane Lafon.
\newblock Diffusion maps.
\newblock {\em Applied and computational harmonic analysis}, 21(1):5--30, 2006.

\bibitem{crump2013evaluating}
Matthew~JC Crump, John~V McDonnell, and Todd~M Gureckis.
\newblock Evaluating amazon's mechanical turk as a tool for experimental
  behavioral research.
\newblock {\em PloS one}, 8(3):e57410, 2013.

\bibitem{cutler2000robust}
Ross Cutler and Larry~S. Davis.
\newblock Robust real-time periodic motion detection, analysis, and
  applications.
\newblock {\em IEEE Transactions on Pattern Analysis and Machine Intelligence},
  22(8):781--796, 2000.

\bibitem{de2002yin}
Alain De~Cheveign{\'e} and Hideki Kawahara.
\newblock Yin, a fundamental frequency estimator for speech and music.
\newblock {\em The Journal of the Acoustical Society of America},
  111(4):1917--1930, 2002.

\bibitem{delbracio2015removing}
Mauricio Delbracio and Guillermo Sapiro.
\newblock Removing camera shake via weighted fourier burst accumulation.
\newblock {\em IEEE Transactions on Image Processing}, 24(11):3293--3307, 2015.

\bibitem{deliyski2007clinical}
Dimitar~D Deliyski, Pencho~P Petrushev, Heather~Shaw Bonilha, Terri~Treman
  Gerlach, Bonnie Martin-Harris, and Robert~E Hillman.
\newblock Clinical implementation of laryngeal high-speed videoendoscopy:
  challenges and evolution.
\newblock {\em Folia Phoniatrica et Logopaedica}, 60(1):33--44, 2007.

\bibitem{goldenberg2005behavior}
Roman Goldenberg, Ron Kimmel, Ehud Rivlin, and Michael Rudzsky.
\newblock Behavior classification by eigendecomposition of periodic motions.
\newblock {\em Pattern Recognition}, 38(7):1033--1043, 2005.

\bibitem{gollub1975onset}
Jerry~P Gollub and Harry~L Swinney.
\newblock Onset of turbulence in a rotating fluid.
\newblock {\em Physical Review Letters}, 35(14):927, 1975.

\bibitem{hatcher2002algebraic}
Allen Hatcher.
\newblock {\em Algebraic topology}.
\newblock University Press Ltd., 2002.

\bibitem{herbst2016phasegram}
Christian~T Herbst, Jakob Unger, Hanspeter Herzel, Jan~G {\v{S}}vec, and
  J{\"o}rg Lohscheller.
\newblock Phasegram analysis of vocal fold vibration documented with laryngeal
  high-speed video endoscopy.
\newblock {\em Journal of Voice}, 30(6):771--e1, 2016.

\bibitem{herzel1994analysis}
Hanspeter Herzel, David Berry, Ingo~R Titze, and Marwa Saleh.
\newblock Analysis of vocal disorders with methods from nonlinear dynamics.
\newblock {\em Journal of Speech, Language, and Hearing Research},
  37(5):1008--1019, 1994.

\bibitem{herzel1996biphonation}
Hanspeter Herzel, Robert Reuter, and Richard~A Katz.
\newblock Biphonation in voice signals.
\newblock In {\em AIP Conference Proceedings}, volume 375, pages 644--657. AIP,
  1996.

\bibitem{huang2010shape}
Peng Huang, Adrian Hilton, and Jonathan Starck.
\newblock Shape similarity for 3d video sequences of people.
\newblock {\em International Journal of Computer Vision}, 89(2-3):362--381,
  2010.

\bibitem{huang2016camera}
Shiyao Huang, Xianghua Ying, Jiangpeng Rong, Zeyu Shang, and Hongbin Zha.
\newblock Camera calibration from periodic motion of a pedestrian.
\newblock In {\em Proceedings of the IEEE Conference on Computer Vision and
  Pattern Recognition}, pages 3025--3033, 2016.

\bibitem{jiang2011statistical}
Xiaoye Jiang, Lek-Heng Lim, Yuan Yao, and Yinyu Ye.
\newblock Statistical ranking and combinatorial hodge theory.
\newblock {\em Mathematical Programming}, 127(1):203--244, 2011.

\bibitem{kantz2004nonlinear}
Holger Kantz and Thomas Schreiber.
\newblock {\em Nonlinear time series analysis}, volume~7.
\newblock Cambridge university press, 2004.

\bibitem{kendall1938new}
Maurice~G Kendall.
\newblock A new measure of rank correlation.
\newblock {\em Biometrika}, 30(1/2):81--93, 1938.

\bibitem{kennel1992determining}
Matthew~B Kennel, Reggie Brown, and Henry~DI Abarbanel.
\newblock Determining embedding dimension for phase-space reconstruction using
  a geometrical construction.
\newblock {\em Physical review A}, 45(6):3403, 1992.

\bibitem{kumdee2015repetitive}
Orrawan Kumdee and Panrasee Ritthipravat.
\newblock Repetitive motion detection for human behavior understanding from
  video images.
\newblock In {\em Signal Processing and Information Technology (ISSPIT), 2015
  IEEE International Symposium on}, pages 484--489. IEEE, 2015.

\bibitem{levy2015live}
Ofir Levy and Lior Wolf.
\newblock Live repetition counting.
\newblock In {\em Proceedings of the IEEE International Conference on Computer
  Vision}, pages 3020--3028, 2015.

\bibitem{lohscheller2007clinically}
J{\"o}rg Lohscheller, Hikmet Toy, Frank Rosanowski, Ulrich Eysholdt, and
  Michael D{\"o}llinger.
\newblock Clinically evaluated procedure for the reconstruction of vocal fold
  vibrations from endoscopic digital high-speed videos.
\newblock {\em Medical image analysis}, 11(4):400--413, 2007.

\bibitem{Mcleod05asmarter}
Philip Mcleod and Geoff Wyvill.
\newblock A smarter way to find pitch.
\newblock In {\em In Proceedings of the International Computer Music Conference
  (ICMC’05}, pages 138--141, 2005.

\bibitem{mehta2011automated}
Daryush~D Mehta, Dimitar~D Deliyski, Thomas~F Quatieri, and Robert~E Hillman.
\newblock Automated measurement of vocal fold vibratory asymmetry from
  high-speed videoendoscopy recordings.
\newblock {\em Journal of Speech, Language, and Hearing Research},
  54(1):47--54, 2011.

\bibitem{miller1956magical}
George~A Miller.
\newblock The magical number seven, plus or minus two: some limits on our
  capacity for processing information.
\newblock {\em Psychological review}, 63(2):81, 1956.

\bibitem{neubauer2001spatio}
J{\"u}rgen Neubauer, Patrick Mergell, Ulrich Eysholdt, and Hanspeter Herzel.
\newblock Spatio-temporal analysis of irregular vocal fold oscillations:
  Biphonation due to desynchronization of spatial modes.
\newblock {\em The Journal of the Acoustical Society of America},
  110(6):3179--3192, 2001.

\bibitem{niyogi1994analyzing}
Sourabh~A Niyogi, Edward~H Adelson, et~al.
\newblock Analyzing and recognizing walking figures in xyt.
\newblock In {\em CVPR}, volume~94, pages 469--474, 1994.

\bibitem{perea2016persistent}
Jose~A Perea.
\newblock Persistent homology of toroidal sliding window embeddings.
\newblock In {\em Acoustics, Speech and Signal Processing (ICASSP), 2016 IEEE
  International Conference on}, pages 6435--6439. IEEE, 2016.

\bibitem{perea2015sliding}
Jose~A Perea and John Harer.
\newblock Sliding windows and persistence: An application of topological
  methods to signal analysis.
\newblock {\em Foundations of Computational Mathematics}, 15(3):799--838, 2015.

\bibitem{pinsky2002introduction}
Mark~A Pinsky.
\newblock {\em Introduction to Fourier analysis and wavelets}, volume 102.
\newblock American Mathematical Soc., 2002.

\bibitem{plotnik2002quantification}
Aaron~M Plotnik and Stephen~M Rock.
\newblock Quantification of cyclic motion of marine animals from computer
  vision.
\newblock In {\em OCEANS'02 MTS/IEEE}, volume~3, pages 1575--1581. IEEE, 2002.

\bibitem{polana1997detection}
Ramprasad Polana and Randal~C Nelson.
\newblock Detection and recognition of periodic, nonrigid motion.
\newblock {\em International Journal of Computer Vision}, 23(3):261--282, 1997.

\bibitem{qiu2003automatic}
Qingjun Qiu, HK~Schutte, Lide Gu, and Qilian Yu.
\newblock An automatic method to quantify the vibration properties of human
  vocal folds via videokymography.
\newblock {\em Folia Phoniatrica et Logopaedica}, 55(3):128--136, 2003.

\bibitem{schuldt2004recognizing}
Christian Schuldt, Ivan Laptev, and Barbara Caputo.
\newblock Recognizing human actions: a local svm approach.
\newblock In {\em Pattern Recognition, 2004. ICPR 2004. Proceedings of the 17th
  International Conference on}, volume~3, pages 32--36. IEEE, 2004.

\bibitem{seitz1997view}
Steven~M Seitz and Charles~R Dyer.
\newblock View-invariant analysis of cyclic motion.
\newblock {\em International Journal of Computer Vision}, 25(3):231--251, 1997.

\bibitem{takens1981detecting}
Floris Takens.
\newblock Detecting strange attractors in turbulence.
\newblock In {\em Dynamical systems and turbulence, Warwick 1980}, pages
  366--381. Springer, 1981.

\bibitem{tralie2016high}
Christopher Tralie.
\newblock High-dimensional geometry of sliding window embeddings of periodic
  videos.
\newblock In {\em LIPIcs-Leibniz International Proceedings in Informatics},
  volume~51. Schloss Dagstuhl-Leibniz-Zentrum fuer Informatik, 2016.

\bibitem{turk1991eigenfaces}
Matthew Turk and Alex Pentland.
\newblock Eigenfaces for recognition.
\newblock {\em Journal of cognitive neuroscience}, 3(1):71--86, 1991.

\bibitem{vejdemo2015cohomological}
Mikael Vejdemo-Johansson, Florian~T Pokorny, Primoz Skraba, and Danica Kragic.
\newblock Cohomological learning of periodic motion.
\newblock {\em Applicable Algebra in Engineering, Communication and Computing},
  26(1-2):5--26, 2015.

\bibitem{venkataraman2016shape}
V~Venkataraman and P~Turaga.
\newblock Shape descriptions of nonlinear dynamical systems for video-based
  inference.
\newblock {\em IEEE transactions on pattern analysis and machine intelligence},
  2016.

\bibitem{wang2009quasi}
Ping Wang, Gregory~D Abowd, and James~M Rehg.
\newblock Quasi-periodic event analysis for social game retrieval.
\newblock In {\em Computer Vision, 2009 IEEE 12th International Conference on},
  pages 112--119. IEEE, 2009.

\bibitem{wilden1998subharmonics}
Inka Wilden, Hanspeter Herzel, Gustav Peters, and G{\"u}nter Tembrock.
\newblock Subharmonics, biphonation, and deterministic chaos in mammal
  vocalization.
\newblock {\em Bioacoustics}, 9(3):171--196, 1998.

\bibitem{wittenberg1995recording}
Thomas Wittenberg, Manfred Moser, Monika Tigges, and Ulrich Eysholdt.
\newblock Recording, processing, and analysis of digital high-speed sequences
  in glottography.
\newblock {\em Machine vision and applications}, 8(6):399--404, 1995.

\bibitem{yaira2016no}
Or~Yair, Ronen Talmon, Ronald~R Coifman, and Ioannis~G Kevrekidis.
\newblock No equations, no parameters, no variables: data, and the
  reconstruction of normal forms by learning informed observation geometries.
\newblock {\em arXiv preprint arXiv:1612.03195}, 2016.

\bibitem{yang2016time}
Jing Yang, Hong Zhang, and Guohua Peng.
\newblock Time-domain period detection in short-duration videos.
\newblock {\em Signal, Image and Video Processing}, 10(4):695--702, 2016.

\bibitem{yu2012solving}
Guoshen Yu, Guillermo Sapiro, and St{\'e}phane Mallat.
\newblock Solving inverse problems with piecewise linear estimators: From
  gaussian mixture models to structured sparsity.
\newblock {\em IEEE Transactions on Image Processing}, 21(5):2481--2499, 2012.

\bibitem{zacharias2016comparison}
Stephanie~RC Zacharias, Charles~M Myer, Jareen Meinzen-Derr, Lisa Kelchner,
  Dimitar~D Deliyski, and Alessandro de~Alarc{\'o}n.
\newblock Comparison of videostroboscopy and high-speed videoendoscopy in
  evaluation of supraglottic phonation.
\newblock {\em Annals of Otology, Rhinology \& Laryngology}, page
  0003489416656205, 2016.

\bibitem{zomorodian05computing}
A.~Zomorodian and G.~Carlsson.
\newblock {Computing persistent homology}.
\newblock {\em Discrete \& Computational Geometry}, 33(2):249--274, 2005.

\end{thebibliography}

\end{document}